\documentclass[conference]{IEEEtran}
\IEEEoverridecommandlockouts

\usepackage{algorithm}
\usepackage{algorithmicx}
\usepackage{algpseudocode}
\usepackage[caption=false]{subfig}
\usepackage{float}
\usepackage{textcomp}
\usepackage{xcolor}
\usepackage{graphicx}
\usepackage{acro}
\usepackage{enumerate}
\usepackage{booktabs}
\usepackage{amsthm}
\usepackage{amsmath,amssymb,amsfonts}
\usepackage{balance}

\newtheorem{lemma}{Lemma}
\usepackage{float}
\usepackage[unicode,hidelinks]{hyperref}
\usepackage{url}

\DeclareAcronym{mtp}{short = MTP, long = Master Thesis Project}
\DeclareAcronym{ldp}{short = LDP, long = Local Differential Privacy}
\DeclareAcronym{fl}{short = FL, long = Federated Learning, short-indefinite = an}
\DeclareAcronym{dp}{short = DP, long = Differential Privacy}
\DeclareAcronym{dnn}{short = DNN, long = Deep Neural Network}
\DeclareAcronym{cnn}{short = CNN, long = Convolutional Neural Network}
\DeclareAcronym{fppdl}{short = FPPDL, long = Fair and Privacy-Preserving Deep Learning}
\DeclareAcronym{dl}{short = DL, long = Deep Learning}
\DeclareAcronym{ldpfl}{short = LDPFL, long = Locally Differentially Private Federated Learning}
\DeclareAcronym{sgd}{short = SGD, long = Stochastic Gradient Descent}
\begin{document}

\title{Strategic Incentivization for Locally Differentially Private Federated Learning}

\author{
    \IEEEauthorblockN{Yashwant Krishna Pagoti\IEEEauthorrefmark{1}, Arunesh Sinha\IEEEauthorrefmark{2}, Shamik Sural\IEEEauthorrefmark{1}}
    \IEEEauthorblockA{
    \IEEEauthorrefmark{1}Indian Institute of Technology Kharagpur, India\\
    \IEEEauthorrefmark{2}Rutgers University, USA\\
    Email: pagotiyashwantkrishna@gmail.com,
    arunesh.sinha@rutgers.edu, shamik@cse.iitkgp.ac.in
    } 
}

\maketitle

\begin{abstract}
In Federated Learning (FL), multiple clients jointly train a machine learning model by sharing gradient information, instead of raw data, with a server over multiple rounds. To address the possibility of information leakage in spite of sharing only the gradients, Local Differential Privacy (LDP) is often used. In LDP, clients add a selective amount of noise to the gradients before sending the same to the server. Although such noise addition protects the privacy of clients, it leads to a degradation in global model accuracy. In this paper, we model this privacy-accuracy trade-off as a game, where the sever incentivizes the clients to add a lower degree of noise for achieving higher accuracy, while the clients attempt to preserve their privacy at the cost of a potential loss in accuracy. A token based incentivization mechanism is introduced in which the quantum of tokens credited to a client in an FL round is a function of the degree of perturbation of its gradients. The client can later access a newly updated global model only after acquiring enough tokens, which are to be deducted from its balance. We identify the players, their actions and payoff, and perform a strategic analysis of the game. Extensive experiments were carried out to study the impact of different parameters. 

\end{abstract}






\maketitle

\section{Introduction}
\label{sec:intro}
\acl{fl} (\ac{fl}) allows multiple clients to train a model by sharing their local gradients with a central server for training over multiple rounds. 
To further prevent data leakage through different forms of inference attacks on FL \cite{towards-efficient-privacy-preserving-fl}, use of Local Differential Privacy (LDP) has been proposed \cite{ding2017collecting}. In Locally Differentially Private Federated Learning (LDP-FL), clients randomly perturb their gradients before sending to the server. 
However, LDP-FL faces a critical challenge in ensuring fair participation while attempting to achieve accuracy of the global model and respecting the privacy concerns of individual clients. The clients tend to contribute differently to the model as their degree of participation varies based on a privacy budget and the perceived value of their contributions.


Thus, there are two opposing factors affecting the success of an LDP-FL set up. The goal of the server is to achieve high global model accuracy and hence, would prefer the least possible perturbation of gradients done by the clients. The clients, on the other hand, are more inclined to behave in a way that protects their privacy and tend to add more noise to their gradients. However, if all the clients overly perturb their gradients, eventually the accuracy of the global model will suffer, rendering the LDP-FL process ineffective. 
This impasse needs to be broken and one way of doing it is to introduce an appropriate incentivization mechanism in the LDP-FL process.

There have been some attempts towards offering monetary incentives to increase participation of clients in LDP-FL \cite{gt1}, \cite{gt2}, \cite{gt3}, \cite{gt4}, \cite{gt5}, \cite{gt6}. However, such an approach does not work universally as clients are often more interested in model performance than payments. In a recent work, Chaudhury et al. \cite{dbsec} introduced a permissioned blockchain-based incentive mechanism for LDP-FL named as BTLF. In a later improvement, which addresses the vulnerabilities of BTLF to adversarial attacks, the same authors proposed a scheme named SBTLF (Secure-BTLF) that cryptographically protects the tokens used as incentives from malicious attacks \cite{saptarshi-IEEETP}. However, even SBTLF works in a non-strategic set up and ignores the conflicting goals of the server and the clients as identified above that can undermine the LDP-FL process itself.  



In this paper, we address the shortcomings of existing work by formulating the LDP-FL process as a game in which the goal of the server is to reach a high global model accuracy and that of the clients is to preserve its privacy even at the cost of some degradation in accuracy. The apparently best possible scenario for an individual client is where it preserves its own privacy while the rest of the clients sacrifice theirs, and yet a high global accuracy is reached. 
However, if all the clients follow such an approach, the FL process will eventually fail. 

Towards our objective, we formulate a token-based incentive scheme in which clients are awarded tokens by the server as and when they participate in an FL round. The number of tokens a client acquires by participating in a given round depends on the level of noise added to its gradients in that round. Higher the quantum of noise, lower is the number of tokens acquired, and vice versa. 
Later, when the client intends to obtain the global model from the server, it has to "buy" the same using its previously acquired tokens. Thus, only if sufficient tokens were accumulated in the past, would a client later be able to access the global model; otherwise, not. A strategic analysis is done to derive the optimal steps to be followed by the clients. 

Note that our approach is different from the Stackelberg game-based modeling of FL done in~\cite{incentive-private-fl-iot}. In that paper, the pricing scheme relies on knowledge of payoffs and cost of clients and the server, which is further specified subjectively by the authors. The scheme also requires actual monetary payments based on these subjective costs. 
We, instead, focus on a mechanism design solution based on artificial currency (tokens). We show that the server can extract the desired privacy level from clients till the point that the client's privacy cost exceeds the incremental value of obtaining a newer learning model, without any information about the payoff and costs of clients. Several experiments were carried out using the MNIST dataset \cite{mnist} and CIFAR10 dataset \cite{cifar10} with varying parameters of the incentive scheme. The level of participation and perturbation strategies of clients as well as model accuracy improvement over multiple FL rounds were studied. Comparative evaluation was done with a non-strategic incentive scheme proposed in~\cite{saptarshi-IEEETP} as the baseline.

\section{Preliminaries}
\label{sec:prelim}
This section introduces some of the foundational concepts required for a deeper understanding of the main contributions made in the paper. It includes FL, LDP and game theory.

\subsection{Federated Learning}
\label{subsec:prelimFL}
\acl{fl} is a machine learning technique that enables multiple entities (clients) to collaboratively train a model while keeping their data local \cite{fedavg}. This is in contrast to traditional approaches, where data is centrally stored and used for training. FL is particularly useful in scenarios where data privacy, communication overhead minimization, and data access rights are of critical concern. In \ac{fl}, local models are trained on local data samples while utilizing a central authority like a server for weight aggregation to create a global model, which in turn is shared with all the clients. The global model is then tested on new data samples \cite{ibm_fl}. It is effective if data has to remain private and data sharing is restricted with any central authority. 
Federated Learning use cases can be broadly categorized into two types, namely, cross-device and cross-silo \cite{adv-open-problems-fl}. Cross-device FL pertains to federated learning where numerous clients - typically IoT or mobile devices, participate in training a model. The clients themselves may not be a future beneficiary of the global model. In the cross-silo setup, on the other hand, there are usually a relatively smaller number of clients, which are also more reliable, e.g., a number of organizations willing to contribute to a global model training. Such clients almost invariably have a stake in the accuracy of the eventually trained global model. In this paper, our application domain is the cross-silo setup of FL.    

\subsection{Local Differential Privacy}
\label{subsec:prelimldp}
First proposed in \cite{10.1007/11681878_14}, Differential Privacy (DP) provides a mathematical foundation for privacy of individual elements in a data collection. The aim is to enable data analysis while not revealing sensitive information about any specific user, thus establishing a strong privacy guarantee for individuals. 
In the basic DP model, which is often denoted as Centralized Differential Privacy (CDP), a trusted aggregator (called a server) manages the sensitive data of all the users. Since it is entrusted with privacy protection, the server adds a calibrated amount of noise to the data while sharing the same with the outside world. The goal of DP is to fudge the contribution of any specific element while preserving the overall accuracy of any analysis that can be done with the data. Note that in DP, the server first needs to collect the raw data from the users so that the aggregated information can be released publicly after appropriate perturbation~\cite{yang2020local}. 
A fundamental assumption in CDP is that the data curator is fully trusted,  which need not necessary hold in all real world situations. 

In order to circumvent this problem, a new mechanism called Local Differential Privacy was proposed \cite{kasiviswanathan2011can}\cite{ding2017collecting}. In LDP, a user perturbs its data (gradients in FL) before sending to the server, thereby providing a stronger privacy guarantee. Since the server has a perturbed version of the data, model training or any form of querying can only be performed on such a perturbed dataset. 
LDP is often specified in terms of what is known as $\epsilon$-Local Differential Privacy~\cite{ding2017collecting}.
A randomized mechanism $\mathcal{M}$ is said to satisfy $\epsilon$-Local Differential Privacy over a data collection $D$ if and only if for any pair of input values $v$ and $v' \in D$ and for any
possible output $S \subseteq Range(\mathcal{M})$, 
\begin{displaymath}
P(\mathcal{M}(v) \in S) \le e^{\epsilon} P(\mathcal{M}(v') \in S)
\end{displaymath}
\indent Here $\epsilon > 0$ is the privacy budget - smaller the value of $\epsilon$, stricter is the protection, but with lower data availability, and vice versa.
It is intuitively obvious that Federated Learning would be a problem naturally suited for LDP and expectedly has been investigated from multiple aspects and application domains \cite{towards-efficient-privacy-preserving-fl}, \cite{lu2020differentially}, \cite{truex2020ldpfed,zhao2021local,samsungLDPFL}. 
%



\subsection{Game Theory}
Game theory is a branch of applied mathematics that studies strategic interactions between rational decision makers. It provides tools for analyzing situations where the outcomes depend on the actions of multiple agents, each with its own preferences and strategies. 
Mechanism design is a sub-area in game theory, where a mechanism designer is able to specify rules and parameters of the games in order to induce desired outcomes. Traditionally, mechanism design has been studied in the context of auctions~\cite{nisan2007algorithmic}, often with money exchanged. In our setting, we employ an artificial currency to extract desired high quality model updates from the clients. Given a game, possibly specified by the mechanism designer, a popular game solution concept is one of Nash equilibrium. Suppose there are $n$ players, with player $i$'s action space being a finite set $A_i$ and its utility dependent on all players' actions. All players move simultaneously, and the players' actions together is called an action profile and denoted by a vector $a \in \prod_{i=1}^n A_i$, where $\prod_{i=1}^n A_i$ is the Cartesian product of all action sets. We use $a_i$ to denote the action of player $i$ in action profile $a$. Then, the utility of player $i$ for an action profile $a$ is $u_i(a)$. A Nash equilibrium is an action profile $a^* \in \prod_i A_i$, such that no player has profitable unilateral deviations, i.e., for all $i$, $u_i(a^*) \geq u_i(a')$ where $a'_j = a^*_j$ for $j \neq i$. In words, the Nash equilibrium is that action profile $a^*$ in which a player cannot gain more utility by changing its own action with the other players actions fixed to their action specified in $a^*$.

\section{Proposed Incentivized LDP-FL Scheme}
\label{sec:proposedldpflsetup}

The incentivized federated learning set up proposed in this paper is described in this section. 
We consider learning of model weights for a multi-class neural classifier as the problem to be addressed in an incentivized cross-silo (Refer to Sub-section \ref{subsec:prelimFL}) FL setting. The methodology, however, is model agnostic and can be used in any other ML task where data is distributed across multiple clients. Our overall federated learning process proceeds as follows:
\begin{enumerate}
    \item \textbf{Initialization:} The server shares the model architecture and other meta data with all clients. Each client initializes a local model with randomly assigned weights.
    \item \textbf{Local Training:} Each client trains its model on local dataset for a specified number of batches. This local training uses gradient descent or its variants.
    \item \textbf{Local Testing:} After local training, each client evaluates the model's performance on both its local dataset and a global test dataset (if made available).
    \item \textbf{Sharing Local Model:} A subset of clients send their model gradients to the server after applying LDP using a chosen value of $\epsilon$.
    \item \textbf{Incentivization:} Server credits tokens to the participating clients based on their respective $\epsilon$ values.
    \item \textbf{Global Model Updating and Distribution:} Server aggregates the local updates (e.g., by averaging) to produce a new global model, and sends it back to the clients for the next round. It also tests model accuracy on a global test set.
    \item \textbf{Local Model Updating:} A subset of clients (having sufficient token balance) read the global model to update their local models. Tokens are deducted for these clients. 
\end{enumerate}
Steps (2) to (7) continue iteratively over multiple rounds until convergence (sufficiently high accuracy on global test set).

In Step (1), the FL process is setup with the server deciding the neural model architecture including the number of hidden layers as well as the number of neurons in each layer. 
In the Step(2), the clients independently train their copy of the model on local data using the hyper parameters determined by the server. As is fairly standard in FL literature, we use SGD for training. After training for a specified number of batches, the clients check model accuracy on their local test dataset (Step 3). Since we are considering cross-silo FL setup among collaborating organizations, all of whom have a stake in the accuracy of the global model, a part of the global test dataset may also be made available to the clients for testing on their local model.  
The next step (Step 4) involves perturbation of local gradients by each client using an LDP mechanism with a chosen value of $\epsilon$. While any of the standard algorithms could be employed, we use the method suggested in \cite{ldp-fl-perturbation} due to its ease of implementation. 
Each client selects its preferred $\epsilon$ based on its privacy requirements and perturbs its model gradients using the LDP mechanism before sending them to the central server. A larger $\epsilon$ implies less noise and hence lower privacy, whereas a smaller $\epsilon$ value results in more noise and higher privacy. 
For the same round, different clients can choose different values of $\epsilon$ and the same client can use a different $\epsilon$ in the different rounds it participates. 

In Step (5), after receiving the gradients from the clients participating in that round, the server credits them tokens commensurate with the value of $\epsilon$. It may be argued that, a client can claim to have used a higher value of $\epsilon$ in order to acquire more number of tokens. However, this is not likely to occur. We consider the clients to behave non-maliciously in this context, since there is no monetary incentive. Further, it will only result in a lower accuracy of the trained global model, which is not strategic for the clients in a cross-silo FL scenario. 
The server next updates the global model by aggregating the local updates (Step 6) and tests its accuracy on a global test dataset. This helps the server to decide whether convergence has been achieved and if so, to terminate the FL process. After testing, the updated weights are now ready to be sent back to the clients. In the final step (Step 7), clients individually request the server to send it the global weights from the current round based on a strategic decision. Tokens are deducted by the server for the clients that read these updated weights from the current round.

\section{Incentive Schemes}
\label{sec:incentiveschemes}
Now that we have a clear understanding of how the proposed LDP-FL scheme operates, 
we address the core challenge. While the server's main goal is to train a robust global model that benefits everyone, clients may prioritize their own privacy and be hesitant to share meaningful updates. To bridge this gap, it is critical to introduce schemes that encourage clients to contribute meaningful updates to the global model while also protecting their own privacy. This leads us to the notion of incentivization. Chaudhury et al. \cite{saptarshi-IEEETP} introduced a token-based incentive scheme to motivate clients by rewarding tokens based on privacy levels as these tokens are used by the clients to buy the global model. This simple non-strategic scheme is first described below.

\textbf{Baseline scheme (Non-strategic)}: We choose the token based incentivization scheme proposed in \cite{saptarshi-IEEETP} as our baseline scheme since it is the most similar to our work. In this setup, a non-strategic approach to token-based incentivization establishes a system where clients can only access updated global model parameters if they hold a sufficient number of tokens. Specifically, clients with a lower $\epsilon$ (high privacy) receive fewer tokens than those with higher $\epsilon$ (lower privacy). Allocation of tokens is a simple linear function, where the token reward $T$ for each client is calculated using Equation \ref{eq:baseline}.
\begin{equation}\label{eq:baseline}
T = 0.5 + \frac{\epsilon - \epsilon_{low}}{2 \times (\epsilon_{high} - \epsilon_{low})}
\end{equation}
where $\epsilon_{low}$ and $\epsilon_{high}$ are the lowest and the highest privacy levels, respectively. In the absence of any payoff maximization strategy, this scheme ends up encouraging clients to contribute less accurate data. We, instead, propose a strategic model as introduced next that better balances the trade-off between privacy and accuracy for clients.

\smallskip

\textbf{Game based mechanism}: In this proposed approach, we address the strategic interaction between the server and clients during each round of FL. It is assumed that all the clients are aware of the performance of the global model. This, for example, can be realized by the server releasing a report of the accuracy of the global model on a global test set. The two parties - clients and server, have distinct goals, leading to a balance between model accuracy and data privacy. The server needs to adopt a tokenization scheme that encourages clients to contribute less noisy data while considering their privacy preferences.
Clients must decide when to participate based on their tokens, privacy level, and the benefits of a well-trained global model, while avoiding ``forced eviction'', where they cannot obtain the latest model updates due to lack of tokens. In this section, we introduce a mechanism design framework based on tokens to model the interactions between the server and clients. We analyze this strategic behavior and identify optimal strategies for both parties, which we present in the following sub-sections. 



\subsection{Players and Actions}
\label{subsec:players&actions}
The game interaction is between a server and multiple clients. The problem is one of mechanism design with the server being the designer. It aims for an outcome where every client uses a privacy level that is more than an acceptable level of $\epsilon_a$. There is a fixed privacy level $\epsilon_{\max}$, and any $\epsilon$ beyond $\epsilon_{\max}$ provides no added benefit in training. Clearly, then $\epsilon_a \leq \epsilon_{max}$. In order to achieve this goal, we design a mechanism based on an artificial currency that we call tokens. There are advantages to our scheme, such as not requiring any knowledge of the cost of privacy of each of the clients and not imposing any real monetary payment based on such unknown costs, which is a stringent requirement in other past work~\cite{incentive-private-fl-iot}. Below we list the actions of the different players.

\begin{itemize}
    \item \textbf{Server: Pricing Scheme.} The server publishes a pricing scheme. One part of this is a price function $f(\cdot)$ that awards $f(\epsilon)$ tokens to a client based its chosen privacy level $\epsilon$. The server also sets a cost of $C$ (in tokens) to be paid by the clients for receiving the global model. The initial global model (at $t=0$) is distributed to the clients free of cost. This pricing scheme specifies rules of the game that the clients participate in.
    \item \textbf{Client: Privacy Level Selection.} Each client's action is to choose an $\epsilon$ level that is same in every round, balancing between privacy and the number of tokens awarded (more tokens for less privacy, i.e., more tokens for higher $\epsilon$). Note that the clients have to commit to an $\epsilon$ at the start, which makes the training process easier for the clients and the server.
\end{itemize}

\subsection{Payoff}
\label{subsec:payoff}
\textbf{Server:} The server has no intrinsic payoff. But, the goal of the server is to ensure the global model is well-trained at the end of FL rounds. If all clients use a privacy level of more than $\epsilon_a$, then the global model is expected to improve its performance in every round. 

\textbf{Client:} We first talk about real monetary cost and reward for clients. Note that, tokens do not count as real cost or reward as they are an artificial currency.  Let the value of global model at round $t$ for client $i$ be $V_i(t)$. A client is already in possession of a global model of value $V$ from a previous round, so the payoff on receiving global model at round $t$ is $V_i(t) - V$, which is $\geq$ 0. The inequality is based on the condition that the global model improves its performance in every round. Further, we assume a diminishing return of training the global model, i.e., $V_i(t+k) - V_i(t)$ is a decreasing function of $t$. The value $V_i(t)$ implicitly depends on $\epsilon$ chosen by all the clients; we do not write it explicitly for ease of notation. However, we assume that $V_i(t)$ is not sensitive to a single client adding less noise, i.e., the value $V_i(t)$ does not change if a single client increases its $\epsilon$. 
Next, the client $i$ incurs a cost of privacy $c_i(\epsilon) \geq 0$ for the choice of $\epsilon$ in any round. For $\epsilon \geq \epsilon_{max}$, the cost is the highest. That means, $c_i(\epsilon_{max}) = c_{max}$ and for $\epsilon \geq \epsilon_{max}$ $c_i(\epsilon) = c_{max}$,  where $c_{max}$ is the maximum value of the cost function. It is natural that the cost function $c(\cdot)$ increases with $\epsilon$.



\subsection{Strategic Analysis}
\label{subsec:strategicanalysis}
We next do a strategic analysis in this sub-section. 

\subsubsection{Incentive Mechanism}
\label{subsubsec:incentivemechanism}
We propose the following mechanism based on an assumption that all clients are chosen to participate in each round, which is typically true when the number of clients is not very large - a likely scenario in cross-silo FL setup.
\begin{enumerate}[i]
\item Freshness Scheme: The server publishes a freshness scheme that has two parts. One is a restriction on clients to perform local computation using a global model from within the last $n$ rounds. Any client at round $t$ is allowed to use a global model (for local computation) that the client possesses from at most $t - n$ rounds in the past. This limit forces clients to obtain and use an updated model. The second part is a token validity requirement. In this, tokens expire in $n$ rounds. Thus, a client needs to use up its tokens (to buy global model), otherwise the tokens earned are wasted.
\item Set $C$ to be any positive integer and a multiple of $n$. Choose $f$ to be any strictly monotonic function and set $f(\epsilon_a) = C/n$.
\end{enumerate}

The above mechanism is only in terms of tokens, with no real monetary payment involved. The clients derive benefit from the global model, and such benefit directly compensates for the cost of privacy. Hence, as shown in the next sub-section, clients only participate till the incremental value gained from the global model exceeds the cost of privacy.

\subsubsection{Mechanism Outcome}
\label{subsubsec:mechanismoutcome}
Next, we analyze the outcome, given the mechanism above. Note that for the proposed mechanism and the dependence of $V_i$ on the choice of $\epsilon$ of all the clients, this is a strategic game with simultaneous moves between clients. However, our mechanism produces a desirable outcome and also makes the mathematical analysis tractable.

\noindent \textbf{Client's Optimal Choice of $\epsilon$}: We can immediately prove the following result.
\begin{lemma}
\label{lemma:clientchoice}
    Any client, if participating, will choose an $\epsilon = \epsilon_a$.
\end{lemma}
\begin{proof} We consider the case of clients participating as mentioned in the lemma statement; note that no participation yields 0 rewards. Thus, participating must yield a positive reward.

We first show that $\epsilon \geq \epsilon_a$. Consider a client choosing $\epsilon < \epsilon_a$, then after the first $n$ rounds, this client will have tokens $< n \cdot C/n = C$, which makes the client unable to buy the global model, i.e. a forced eviction. After this, the client is force evicted as the client cannot participate with a global model older than $n$ rounds and it cannot buy a new global model. Thus, the client gains no value (as it does not receive any updated global model) and only incurs privacy cost. Hence, $\epsilon < \epsilon_a$ is not a rational choice for a participating client, since it yields a negative reward.
    
Next, recall that a Nash equilibrium is a set of actions by all clients such that any unilateral deviation by one client reduces the rewards for that client. We show that $\epsilon_a$ for all clients is a Nash equilibrium among the clients. This can be seen by observing that if any one client $i$ unilaterally chooses $\epsilon > \epsilon_a$ (note, $\epsilon < \epsilon_a$ is not rational as per Lemma \ref{lemma:clientchoice}), then it pays more privacy cost $c_i(\epsilon) > c_i(\epsilon_a)$ with no added benefit from the value of the global model (recall that a single client cannot affect model value $V_i$ by increasing its $\epsilon$). Thus, unilateral deviation from $\epsilon_a$ for any client increases the cost of that client with no gain in the value from global model, and hence $\epsilon_a$ for all clients is a Nash equilibrium.
\end{proof}

\noindent \textbf{Server's Optimization Problem}: The server needs to choose $\epsilon_a$ and $n$. Towards this, observe that at the start of round $t$, where $t$ is a multiple of $n$, all clients have 0 tokens and will participate only if
\begin{equation}\label{eq:IR}
    V_i(t +n) - V_i(t) \geq n c_i(\epsilon_a) 
\end{equation}
The RHS above is the real cost (not tokens) of privacy over $n$ rounds and the LHS is the value gained from the improved model over $n$ rounds. The longer the clients participate, the closer the server will reach towards its goal of producing a well-trained global model. To make the clients participate longer, the server must try to reduce the RHS and increase LHS of Inequality \ref{eq:IR}. Thus, $c_i(\epsilon_a)$ should be reduced, which implies $\epsilon_a$ must be reduced (since $c_i$ is monotonic). Hence, the server needs to find a minimum acceptable $\epsilon_a$ that does not cause the training to collapse. This can be done empirically, as we show in our experiments.

Next, note that $(V_i(t +n) - V_t(t)) /n $ is a decreasing function in $n$ because of the diminishing return property of $V_i$. Thus, the best choice of $n$ is $1$. With this, the client participates if
\begin{equation}\label{eq:IRn1}
    V_i(t +1) - V_t(t) \geq  c_i(\epsilon_a) 
\end{equation}
Based on Inequality~\ref{eq:IRn1}, we modify our mechanism appropriately for some cases to obtain better outcomes. 

\subsubsection{Improved Mechanism}
\label{subsubsec:improvedmechanism}
With an aim to increase the LHS of Inequality~\ref{eq:IR}, we propose a mechanism in which \emph{not} all clients are chosen to participate in each round. Instead, the clients are partitioned into a number of groups, say $G$. In each round, one group is chosen to provide model updates. We modify our freshness scheme to take into account the last time a client participated. The changed parts below are italicized.\\
\textbf{Freshness Scheme:} The server publishes a freshness scheme that has two parts. One is a restriction on clients to perform local computation using global models from within the last $n$ rounds \emph{that it participated in}. Any client at round $t$ is allowed to use a global model (for local computation) that the client possesses from at most $t - n$ rounds in the past \emph{that it participated in}. This limit forces clients to obtain and use an updated model \emph{after participating in $n$ rounds}. 
    
For validity requirement, tokens expire \emph{after $n$ rounds that the client participates in}. Thus, a client needs to use tokens (to buy global model), otherwise the tokens earned are wasted.

Lemma~\ref{lemma:clientchoice} still holds with the above changed freshness scheme. Further, the argument for choosing $n=1$ still holds, and $\epsilon_a$ also can still be chosen empirically. The beneficial change is in the inequality in Equation~\ref{eq:IRn1}, which changes to
\begin{equation}
    V_i(t + G) - V_t(t) \geq  c_i(\epsilon_a) 
\end{equation}
The above change is because a client participates after $G$ rounds, thus $t+1$ in Inequality~\ref{eq:IRn1} is replaced by $t+G$. This change makes the LHS of the equation larger, thereby leading to the client willing to participate for more rounds.


\section{Experimental Results}
\label{sec:experiments}

We next present the results of our experiments for which 
we used a virtual machine equipped with an Intel Xeon processor having 32 VCPUs, 64 GB of RAM.
Studies were carried out on two benchmark datasets, namely, MNIST and CIFAR10. While the results of MNIST are presented here, those for CIFAR10 are given in the Appendix.

\subsection{Dataset and Federated Learning Setup}
\label{subsec:exp-fedsetup}
The MNIST dataset is comprised of 70,000 grayscale images of handwritten digits ranging from 0 to 9, with 60,000 images used for training and 10,000 for testing. Each image is of size 28$\times$28 pixels.
For this study, we implemented a neural network model with the following details:

\begin{enumerate}
    \item The input layer flattens the 28$\times$28 pixel images into a single vector of 784 features.
    \item A dense (fully connected) layer with 128 neurons was added, using the ReLU activation function.
    \item The output layer consists of 10 neurons, corresponding to the 10-digit classes in the MNIST dataset.
\end{enumerate}

For model training, we use \ac{sgd} as the optimization algorithm. In the federated learning process, each client trains its local model using a fixed number of batches per round, and the global model is updated after every round of training. The batch size and the number of training rounds were carefully selected to ensure that the model converges effectively within a finite number of rounds.

The training process follows the Federated Averaging (FedAvg) algorithm \cite{fedavg}. In each round of FedAvg, these clients compute the gradients of the loss function using their local datasets. The server aggregates these gradients to update the global model. The update rule for the server is:

\begin{equation}\label{eq:updateruleserver}
    w^{t+1} = w^{t} - \eta \sum_{k=1}^{K} \frac{n_k}{n}g_k
\end{equation}
where $w^t$ is the model weight at round $t$, $\eta$ is the learning rate, $n_k$ is the number of data samples at client $k$, and $g_k$ is the gradient computed by client $k$.


\begin{figure*}
        \centering
	\captionsetup[subfigure]{}
		\subfloat[3 Clients: $\epsilon_{i}$ = (25, 15, 1)
        \label{fig:baseline-acc-c3}]{
			\includegraphics[width=0.45\textwidth,keepaspectratio]{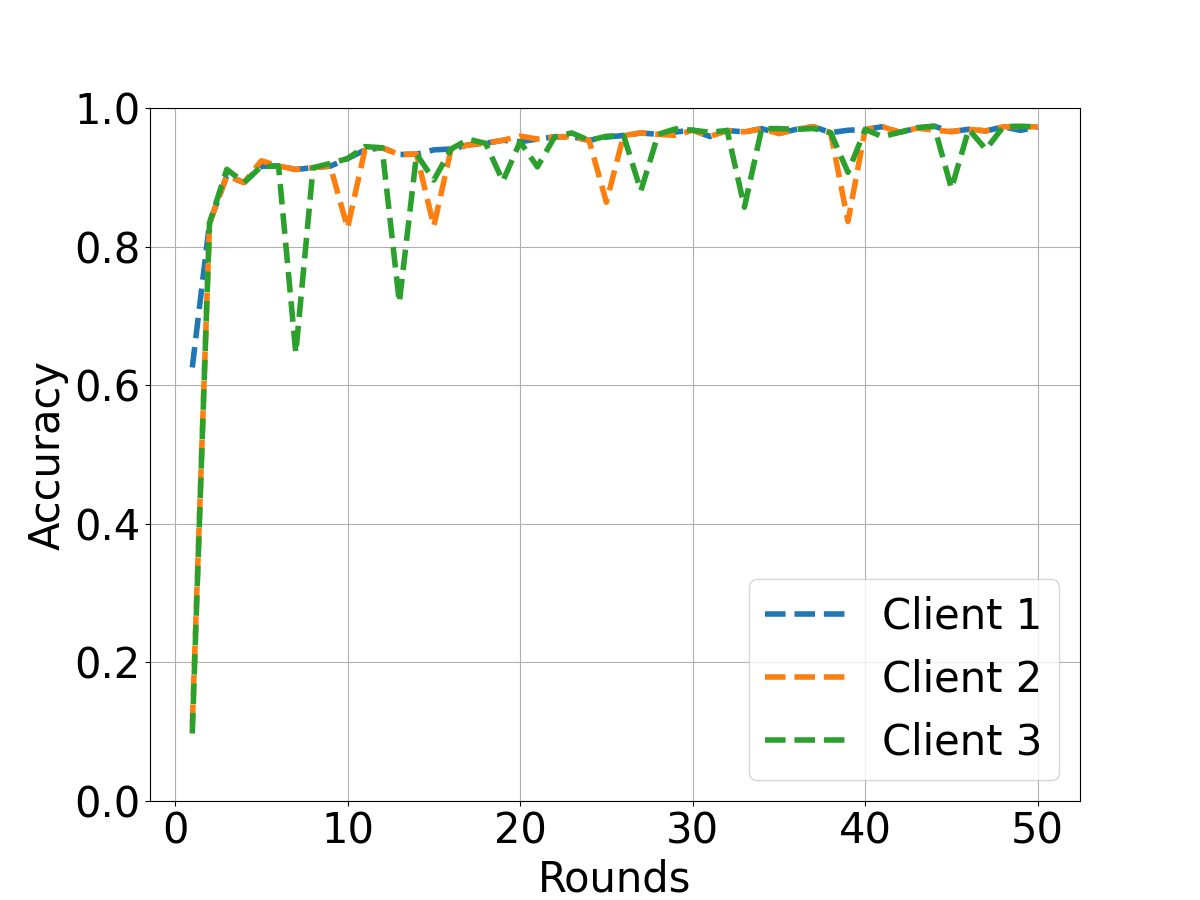}
		}
		\subfloat[10 Clients: $\epsilon_{i}$ = (25, 23, 20, 17, 15, 13, 10, 7, 5, 1)
        \label{fig:baseline-acc-c10}]{
			\includegraphics[width=0.45\textwidth,keepaspectratio]{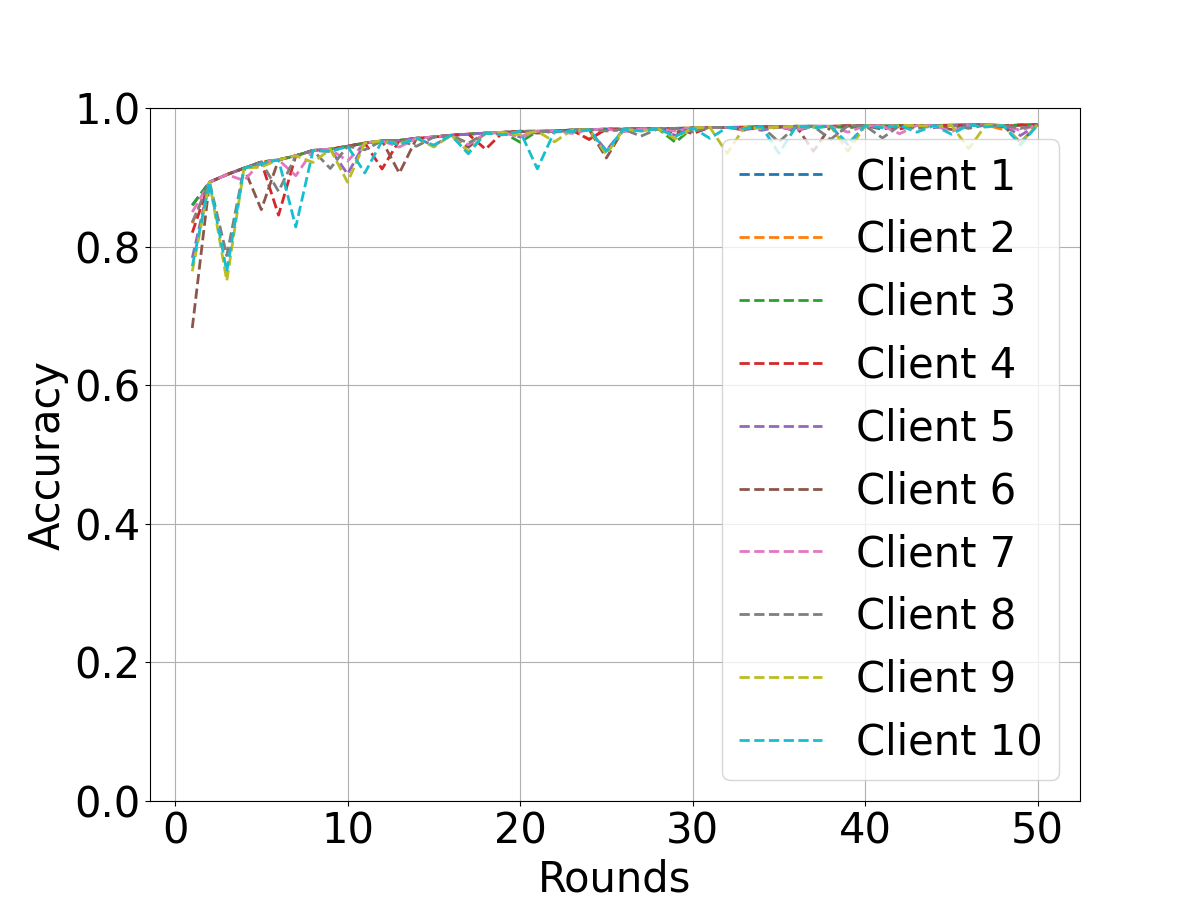}
		}
          \setlength{\belowcaptionskip}{-8pt}
	\caption{Federated Learning Accuracy with Non-Strategic (baseline) Incentive Mechanism SBTLF under Intermediary Data Distribution, tested on the Global Dataset. Accuracies of Clients are shown for (a) 3 clients and (b) 10 clients.}
	\label{fig:baseline-fl}
\end{figure*}

\subsection{Data Distribution among Clients}
\label{subsec:exp-datadistribution}
In \ac{fl}, one of the key aspects to consider is how the data is distributed among the clients, as this impacts model performance during training. 
To address this, we tested different approaches to data distribution. At one extreme, there is \textbf{identical distribution}, which ensures every client gets a uniform sample of the entire dataset. 
At the other extreme, \textbf{disjoint distribution} divides the dataset into distinct parts, with each client receiving data related to specific classes. For instance, in a handwritten digit recognition task from the MNIST dataset, one client might receive only samples of the digit "3", while another gets only samples of "7", and so on. 
It was observed in our experiments that such a scenario gives better result when fewer clients are involved. But, as the number of clients increases, the results start fluctuating and become less reliable.
Considering a more realistic situation in between the two extremes, we introduced an \textbf{intermediary distribution} method. Here, half of the dataset is distributed identically across clients, ensuring a shared representation of general patterns, while the other half is distributed disjointedly. 
The above approach helped to refine strategies to improve the performance of the global model while addressing challenges like non-representative and unbalanced client data \cite{fedavg}.




\subsection{Baseline using Non-strategic Tokenization}
\label{subsec:exp_nonstrategic}
SBTLF, the non-strategic token-based incentive scheme as proposed in \cite{saptarshi-IEEETP} and detailed in Section~\ref{sec:incentiveschemes}, is considered as the baseline for comparison since it is the one which is most similar to our work. With this baseline approach, we perform experiments with varying client privacy levels $\epsilon$ using the dataset with intermediary distribution as mentioned in Sub-section \ref{subsec:exp-datadistribution}.
The intermediary dataset was used since it was found to yield better results for global testing and more closely mimics real-world scenarios. We calculate both the accuracy of the clients and their participation count over the entire LDP-FL process to better understand the effectiveness of the SBTLF mechanism. We tested with three clients as well as as ten clients taking part in the FL process using the MNIST dataset as done in \cite{saptarshi-IEEETP}.

\begin{figure*}[t]
        \centering
	\captionsetup[subfigure]{}
		\subfloat[3 Clients: $\epsilon_{i}$ = (25, 15, 1)
        \label{fig:baseline-pc-c3}]{
			\includegraphics[width=0.45\textwidth,keepaspectratio]{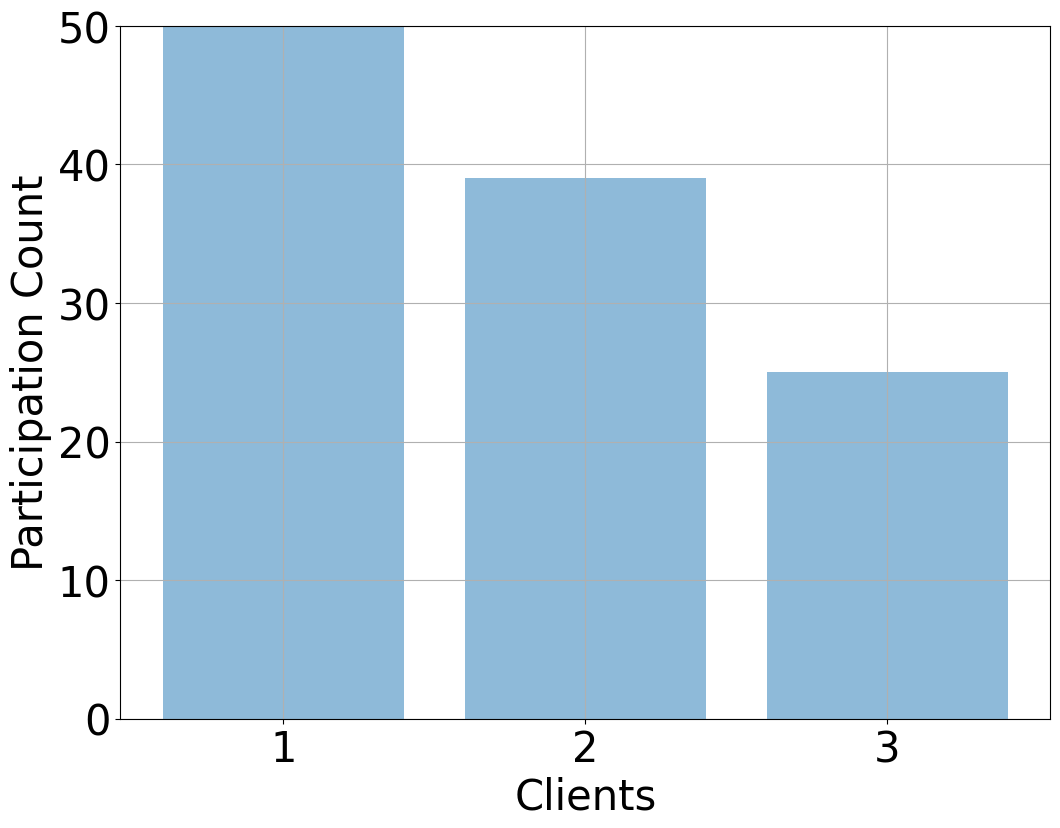}
		}
		\subfloat[10 Clients: $\epsilon_{i}$ = (25, 23, 20, 17, 15, 13, 10, 7, 5, 1)
        \label{fig:baseline-pc-c10}]{
			\includegraphics[width=0.45\textwidth,keepaspectratio]{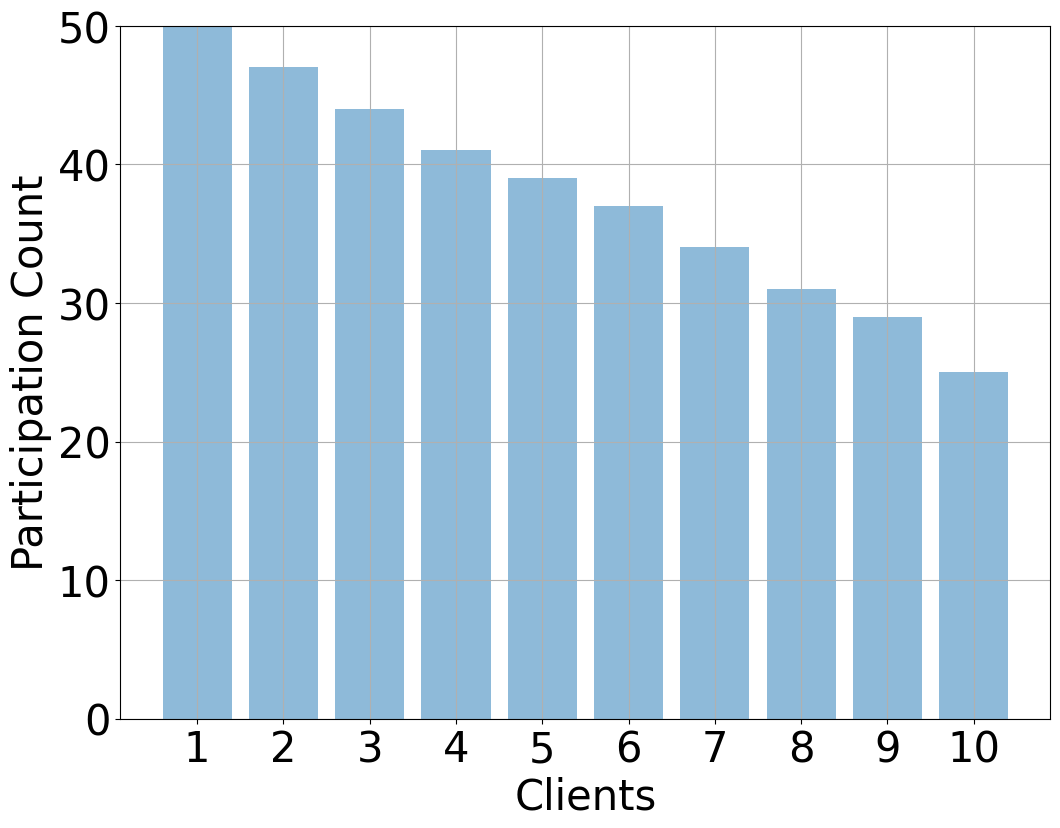}
		}
          \setlength{\belowcaptionskip}{-8pt}
	\caption{Client participation count across rounds in \ac{fl} with Non-Strategic (baseline) Incentive Mechanism SBTLF under intermediary data distribution, tested on the Global Dataset.}
	\label{fig:baseline-pc}
\end{figure*}

\textbf{Three Clients:} In this setup, we consider three clients with $\epsilon$ values 1, 15, and 25. Figure~\ref{fig:baseline-acc-c3} shows how the accuracy of each client changes over rounds. On the x-axis we have the round number, and on the y-axis we have the accuracy of each client. Each client has a different privacy level $\epsilon$, which affects how often they can get access to the global model. In the plot, it is observed that the client with the highest privacy level ($\epsilon$ = 25) continuously receives the global model in every round. However, the other clients, who have lower privacy levels, show a fluctuating accuracy curve as they can only afford the global model at certain rounds when they have enough tokens. Only at those points, accuracies match that of the first client.

\textbf{Ten Clients:} For this experiment, we consider ten clients, each with a distinct $\epsilon$ value: 25, 23, 20, 17, 15, 13, 10, 7, 5, and 1. Figure~\ref{fig:baseline-acc-c10} illustrates the previous experiment, but with more number of clients and varying $\epsilon$ values.

The participation count graphs for the experiment with three clients (Figure~\ref{fig:baseline-pc-c3}) and with ten clients (Figure~\ref{fig:baseline-pc-c10}) further illustrate how many times the clients received the global model according to their levels of privacy. 

\subsection{Results for the Proposed Strategic Incentivization Scheme}
\label{subsec:strategicresults}
In Section \ref{sec:incentiveschemes}, we introduced the need for three functions, namely, Value of the global model $V(\cdot)$, Cost $c(\cdot)$ and Reward $f(\cdot)$. For carrying out experiments, these functions need to be defined. We chose the value and cost functions empirically. A concave function (Equation~\ref{eq:valuefunction}) was chosen as the value function to achieve the diminishing returns property as the rounds progress.
\begin{equation}\label{eq:valuefunction}
V(n) = \frac{30 \times (\log(n + 1))^{2.8}}{1 + 0.15 \times (\log(n + 1))^{1.5}}
\end{equation}
where $n$ is the current round number.

The cost function is a convex function that incurs more cost to the client for higher $\epsilon$. Equation \ref{eq:costfn} represents the cost function.
\begin{equation}
\label{eq:costfn}
    \!\! c(\epsilon) \! =
\begin{cases}
    c_{max}, &\text{if }\epsilon \geq \epsilon_{max} \\ 
    (c_{max} - c_{min}) (\frac{\epsilon - 1}{\epsilon_{max} - 1})^3 + c_{min},  &\text{otherwise}
\end{cases}
\end{equation}
Here $c_{max}$ and $c_{min}$ are respectively the maximum and the minimum costs incurred by a client to participate in a round, whereas $\epsilon_{max}$ is the maximum possible $\epsilon$ value chosen by the client. The function is designed to clamp the cost within a fixed range, ensuring that it stays between $c_{min}$ and $c_{max}$.

The server chooses an acceptable $\epsilon_{a}$. We illustrate a reward function that is convex in nature, to penalize the clients more strictly than that of a function that varies linearly.
\begin{equation}
f(\epsilon) = 
\begin{cases} 
    C, & \text{if } \epsilon \geq \epsilon_{\text{a}} \\
    0.5 + \left(C - 0.5 \right) \left( \frac{\epsilon - \epsilon_{\text{min}}}{\epsilon_{\text{a}} - \epsilon_{\text{min}}} \right)^3, & \text{otherwise}
\end{cases}
\end{equation}
where $e_{min}$ and $e_{max}$ are the minimum and maximum permissible values of the privacy levels, respectively. The reason for setting limits on the privacy levels is to focus on a specific range of privacy values so that we can study a finite set of cases and draw meaningful conclusions.
The functions were chosen so as to be able to visualize how utility changes over the training rounds of participation. For example, if a client selects a privacy level near $\epsilon_a$, it may struggle to participate because either the utility goes below zero (if the chosen $\epsilon > \epsilon_a$), or the clients are not rewarded enough tokens to be able to participate further in the training (if the chosen $\epsilon < \epsilon_a$).

\begin{figure}[t]
    \centering
    \includegraphics[width=0.9\columnwidth]{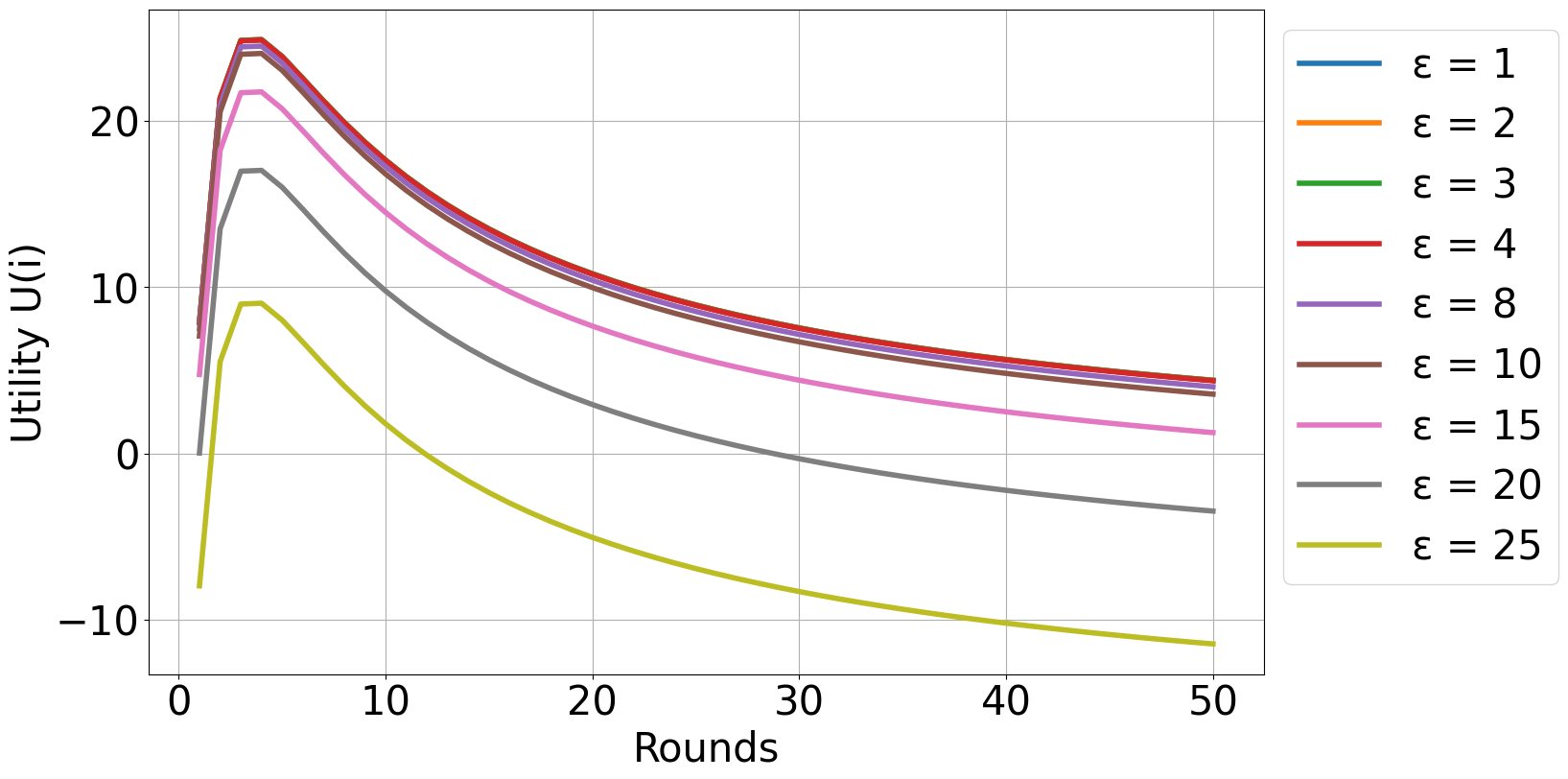}
      \setlength{\belowcaptionskip}{-8pt}
    \caption{Utility vs. No. of rounds for various $\epsilon$ values.} 
    \label{fig:utility}
\end{figure}

\begin{figure*}
  \centering
  \captionsetup[subfigure]{}
  \subfloat[$\epsilon = 25$\label{fig:nmul-c3-max}]{
    \includegraphics[width=0.32\textwidth,keepaspectratio]{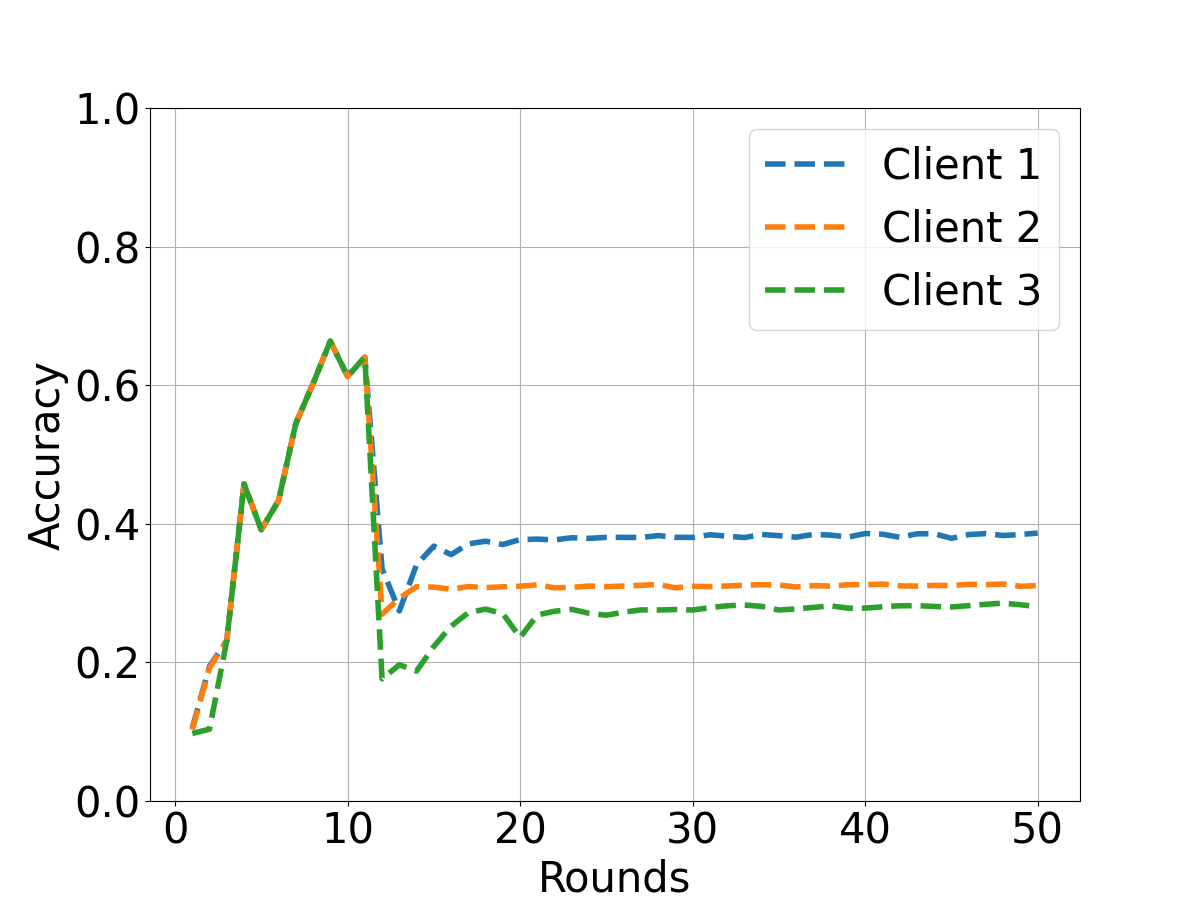}
  }
  \subfloat[$\epsilon = 17$\label{fig:nmul-c3-min}]{
    \includegraphics[width=0.32\textwidth,keepaspectratio]{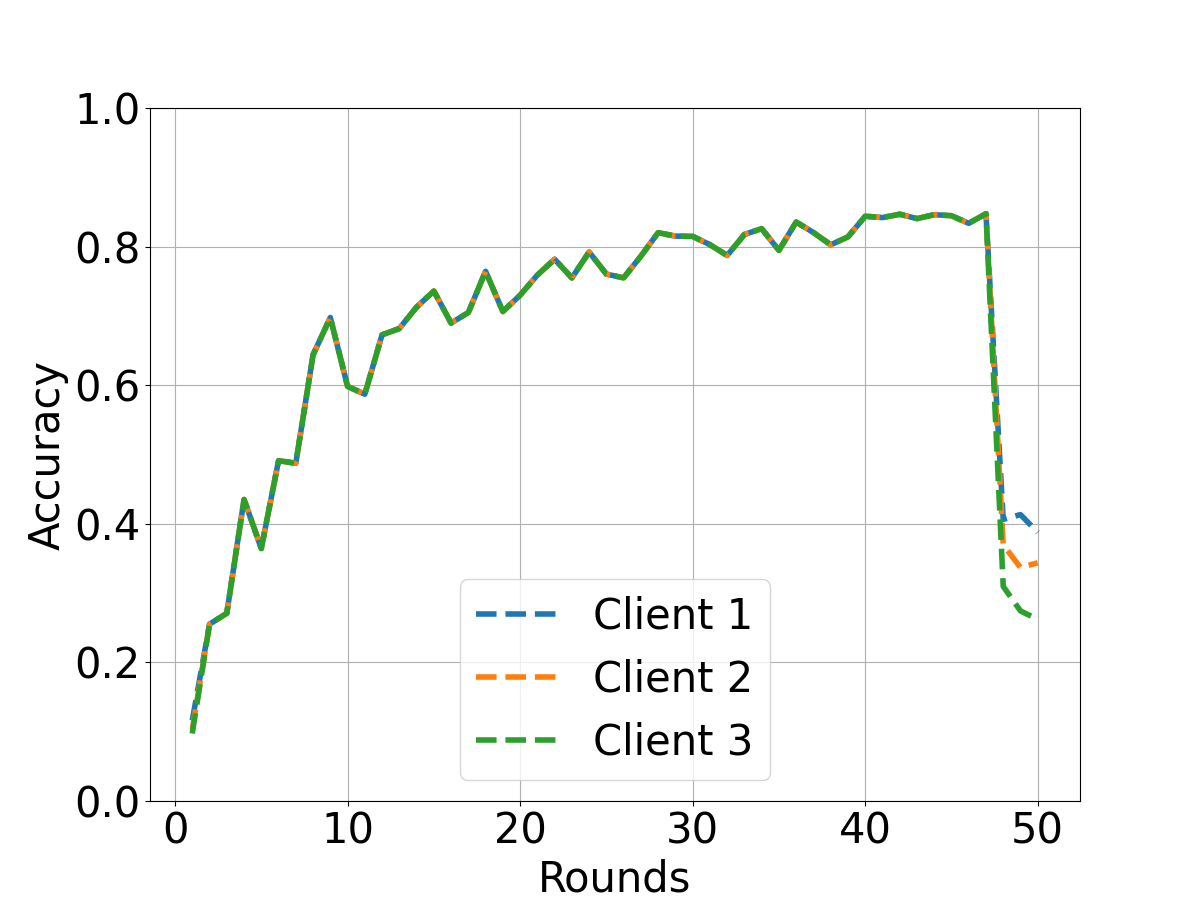}
  }
  \subfloat[$\epsilon = 15$\label{fig:nmul-c3-no}]{
    \includegraphics[width=0.32\textwidth,keepaspectratio]{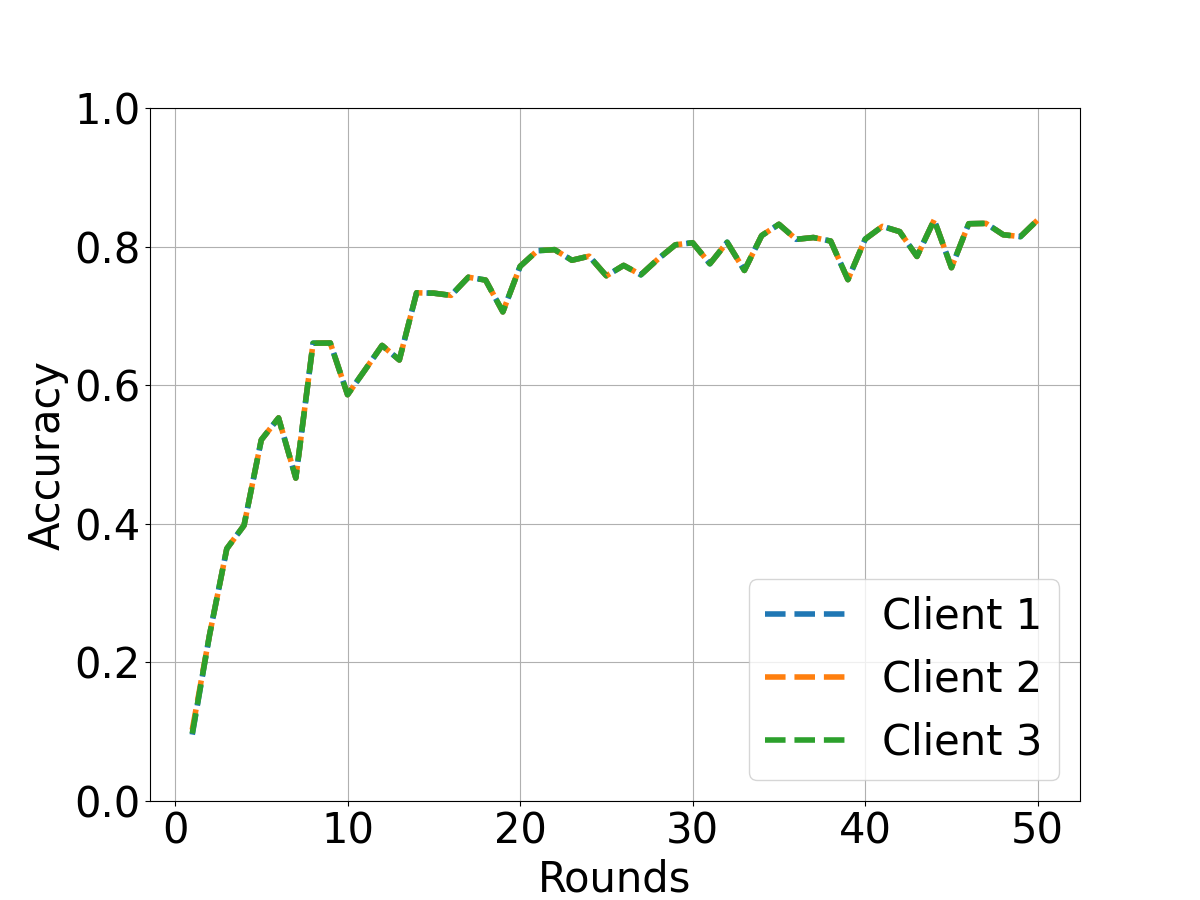}
  }
    \setlength{\belowcaptionskip}{-8pt}
  \caption{FL Accuracy variation for the \textbf{MNIST dataset} tested on a Global Dataset with Disjoint Distribution for \textbf{3} clients illustrating where the training collapse occurs (a) earliest, (b) after most delay, and (c) does not collapse at all.}
  \label{fig:nmul-c3-breakpoint}
\end{figure*}

Now that the functions $V(\cdot)$, $c(\cdot)$, and $f(\cdot)$ are defined, we compute the client's utility as defined in Equation (\ref{eq:IRn1}). The server uses this utility function to evaluate different $\epsilon$ values and determine an appropriate threshold for participation. Figure~\ref{fig:utility} presents utility plotted against the number of rounds for different $\epsilon$ values, independent of the dataset used. Each curve represents a client participating with a different privacy level.
From the graph, it can be observed that when $\epsilon = 15$, the utility remains positive throughout all the rounds. This indicates that at this threshold privacy level, the clients can sustain participation by earning enough tokens to continue acquiring the global model. For $\epsilon = 20$, the utility drops below zero after a few rounds, rendering the clients ineligible to participate further in the process.
For  $\epsilon > 20$, the utility curve drops into the negative region early in the training. This means that clients are penalized for participating with lower privacy levels. On the other hand, when $\epsilon < 15$, the utility remains positive throughout the training, but the clients do not receive sufficient tokens to buy the global model consistently. These clients eventually fall out of the training process as they cannot afford continued participation.

We next examine at which $\epsilon$ values the training collapses. For this, we set the initial $\epsilon$ value same for all clients and perform the training with varying number of clients: 3 (Figure~\ref{fig:nmul-c3-breakpoint}) and 10 (Figure~\ref{fig:nmul-c10-breakpoint}). The figures present the accuracy graph plotted against rounds with the participating clients of 3 and 10, respectively using MNIST. All the clients have a privacy level fixed to the same value at the beginning of the training. In both datasets, we highlight three cases: (a) the maximum $\epsilon$ value where the training collapses, (b) the minimum $\epsilon$ above acceptable
$\epsilon_{a}$ where the training collapses, and (c) where the training continues without collapsing. The $\epsilon$ values for the three cases are: 25, 17 and 15, respectively.

\subsubsection{Three Clients}
In Figure~\ref{fig:nmul-c3-max}, 
we see that the clients stop participating after round 10. This happens because their utility becomes negative, meaning they are no longer benefiting from training. Since they stopped participating, they do not receive the global model updates, which causes their accuracies to spike down abruptly as the clients' training is done on disjoint datasets, while the inference is done on the entire test dataset. As long as the clients receive the aggregated weights from the server, the accuracy will see stable growth until convergence. In Figure~\ref{fig:nmul-c3-min}, a similar trend is observed, except that the utility remains positive longer, until round 42. At this point, the utility becomes negative, leading to a sharp drop in accuracy for all clients as observed in the previous case. 
On the other hand, in Figure~\ref{fig:nmul-c3-no}, 
the participation continues throughout all training rounds, indicating that the incentive mechanism maintains positive utility for the clients. As a result, all clients consistently receive the global model updates, leading to a stable and improving accuracy trend across all rounds.


\subsubsection{Ten Clients}
\begin{figure*}[t]
  \centering
  \captionsetup[subfigure]{}
  \subfloat[$\epsilon = 25$
    \label{fig:nmul-c10-max}]{
    \includegraphics[width=0.32\textwidth,keepaspectratio]{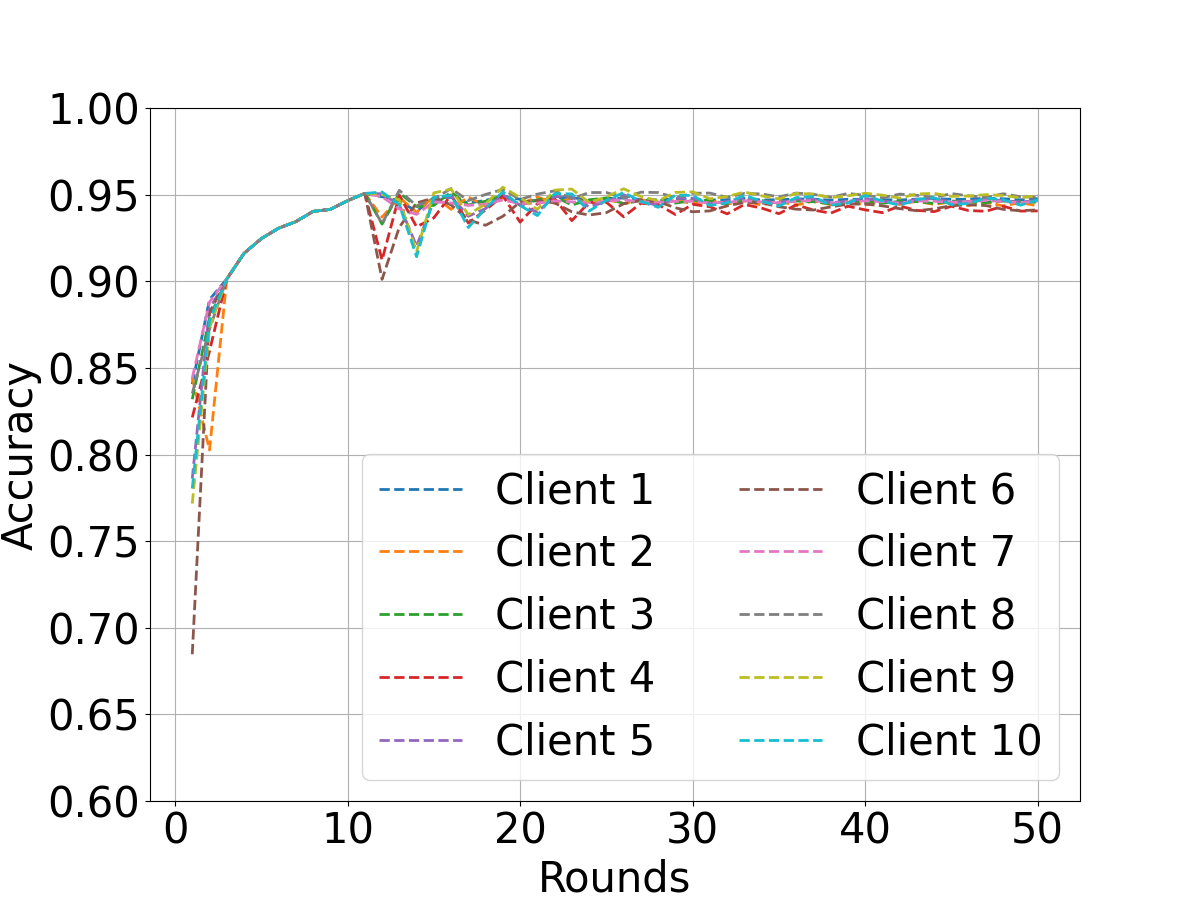}
  }
  \subfloat[$\epsilon = 17$
    \label{fig:nmul-c10-min}]{
    \includegraphics[width=0.32\textwidth,keepaspectratio]{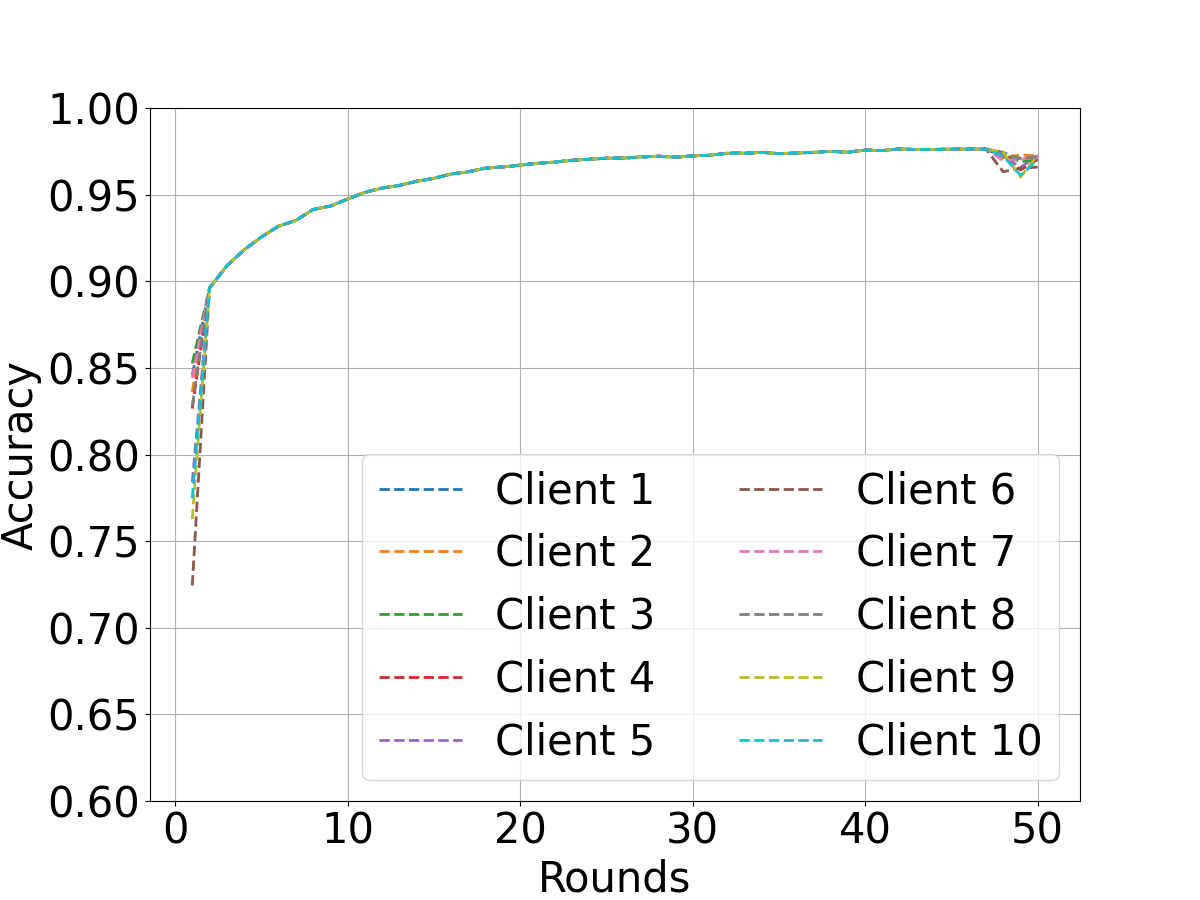}
  }
  \subfloat[$\epsilon = 15$
    \label{fig:nmul-c10-no}]{
    \includegraphics[width=0.32\textwidth,keepaspectratio]{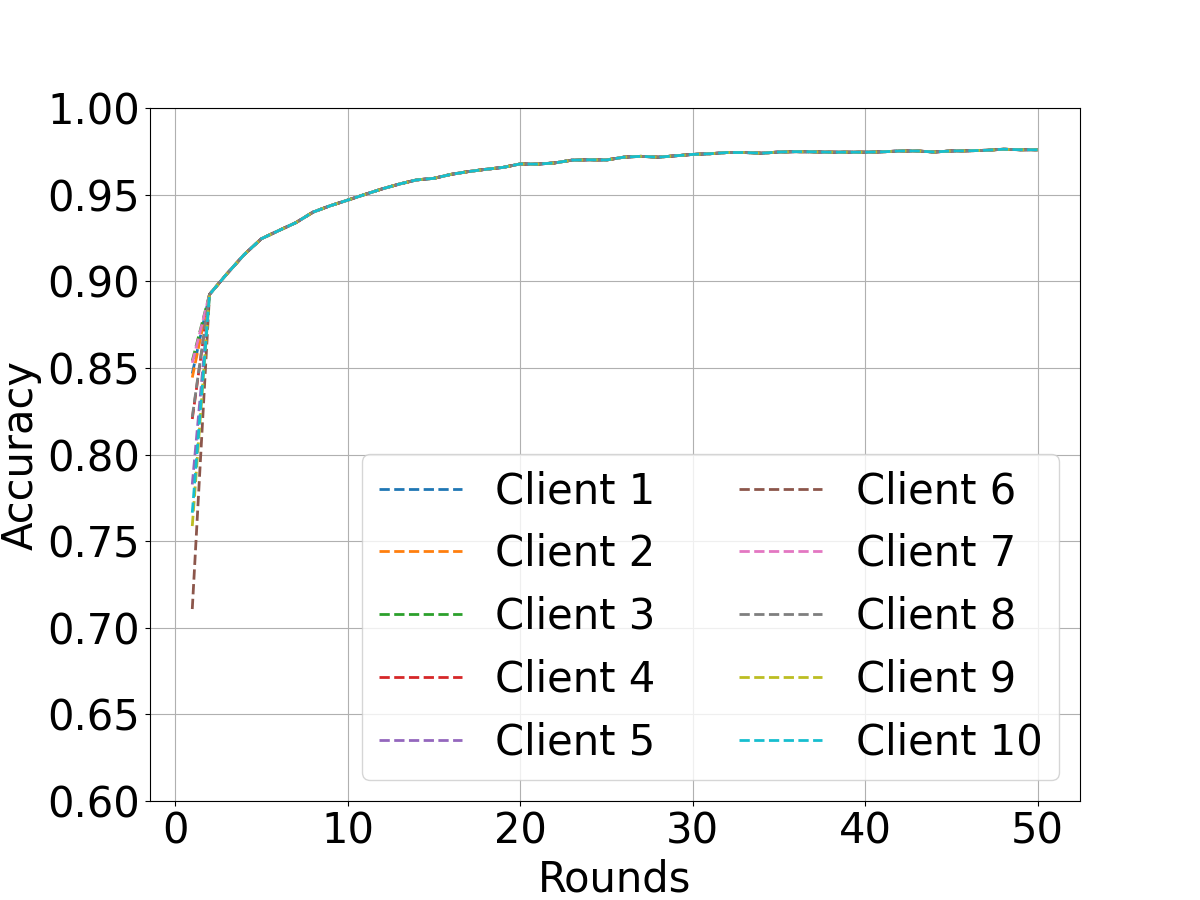}
  }
    \setlength{\belowcaptionskip}{-8pt}
    \caption{FL Accuracy variation for the \textbf{MNIST dataset} tested on a Global Dataset with Disjoint Distribution for \textbf{10} clients illustrating where the training collapse occurs (a) earliest, (b) after most delay, and (c) does not collapse at all.}
  \label{fig:nmul-c10-breakpoint}
\end{figure*}

In Figure~\ref{fig:nmul-c10-max}, 
a pattern in line with the three-client case is observed. The clients stop participating at a certain round, and from that point onward, there is a very slight decrease in accuracy. Additionally, we can see a small spread in accuracy values among clients, indicating that they are not receiving the global model updates after dropping out of training. A similar trend is also observed in Figure~\ref{fig:nmul-c10-min},
but this time, the spread in accuracy appears near the end of training. In contrast, in Figure~\ref{fig:nmul-c10-no}, 
the accuracy curves for all clients stay closely aligned throughout the entire training process. This shows that all of them continued participating and received the global model updates in every round, ensuring stable learning without divergence.

\begin{figure}[t]
  \centering
  \captionsetup[subfigure]{}
  \subfloat[
    \label{fig:nmul-c3-break-pc}]{
    \includegraphics[width=0.8\columnwidth,keepaspectratio]{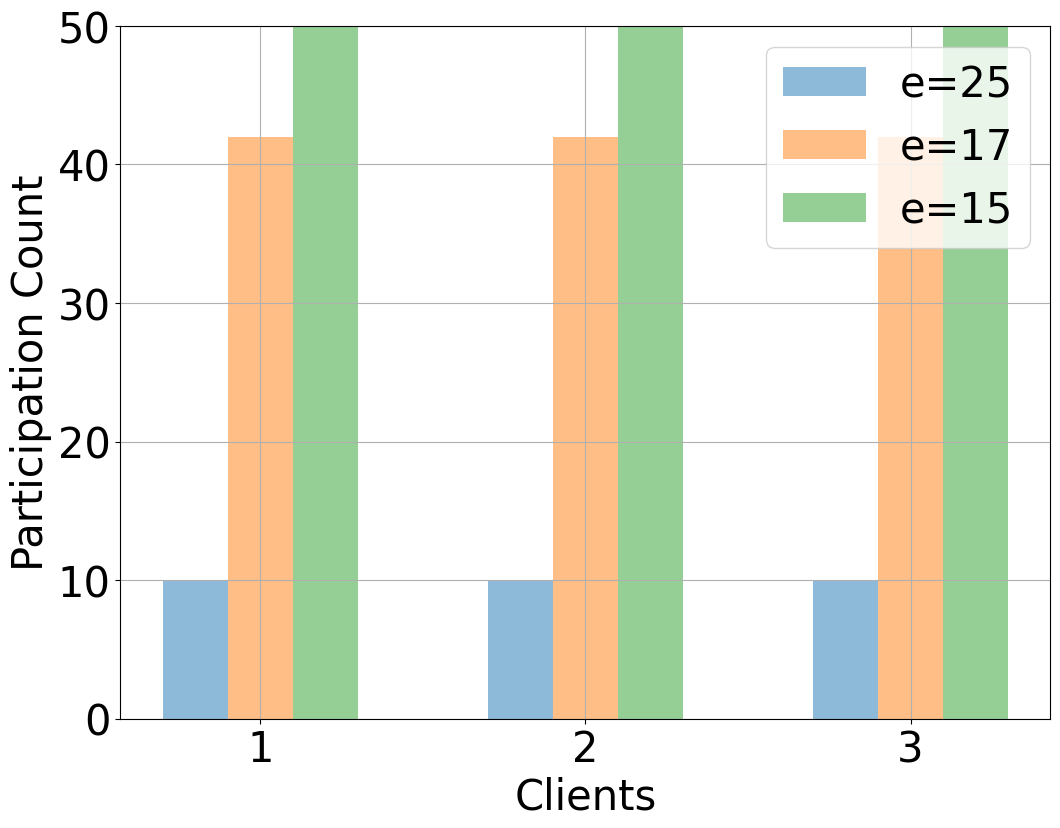}
  }
  \\
  \subfloat[
    \label{fig:nmul-c10-break-pc}]{
    \includegraphics[width=0.8\columnwidth,keepaspectratio]{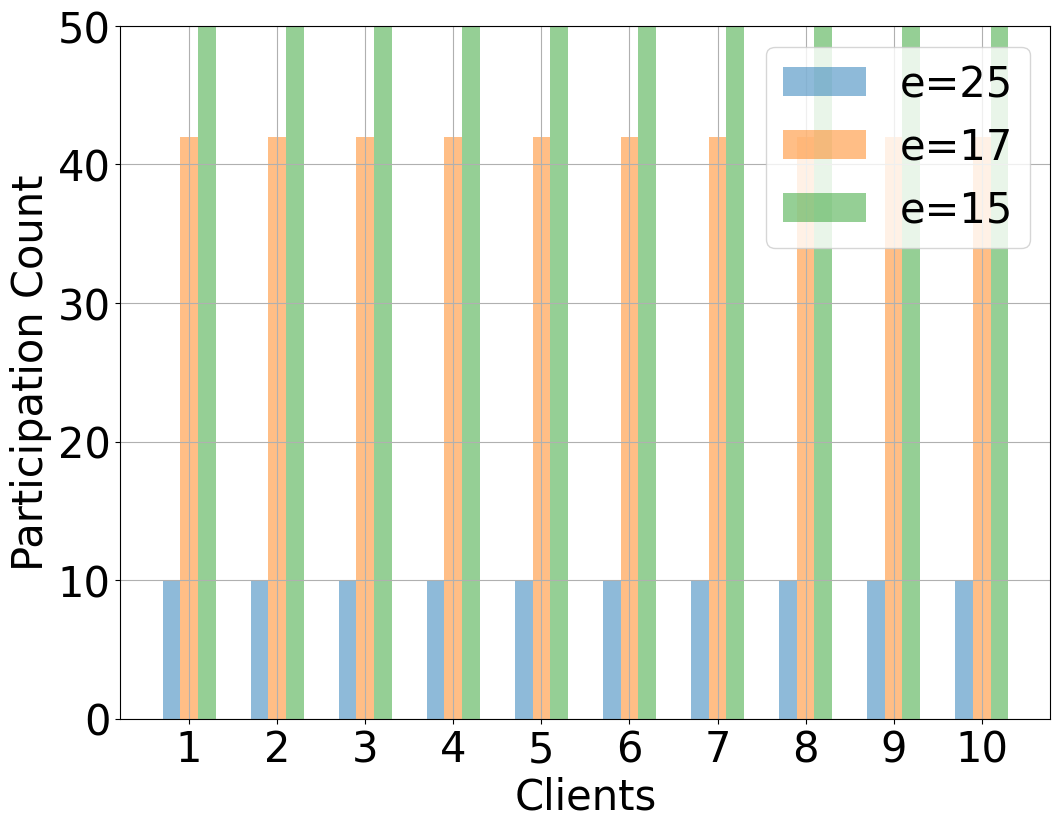}
  }
    \setlength{\belowcaptionskip}{-8pt}
  \caption{Client participation count tested on a Global Dataset using \textbf{MNIST dataset}. The figure illustrates participation across (a) 3 and (b) 10 clients, where the three bars indicate the number of times the Client participated in the cases where training collapse occurs earliest (blue), after most delay (orange), and does not collapse at all (green).}
  \label{fig:nmul-breakpoint-pc}
\end{figure}

Finally, the participation count plots in Figure~\ref{fig:nmul-breakpoint-pc} show how many training rounds each client participated in for the three-client case (Figure~\ref{fig:nmul-c3-break-pc}) and the ten-client case (Figure~\ref{fig:nmul-c10-break-pc}) under different values of $\epsilon$ (25, 17, 15). These trends align with the accuracy behavior observed earlier.
The accuracies of the participating clients tested on the global test dataset are plotted in Figure~\ref{fig:nmul-c3-no}
and Figure~\ref{fig:nmul-c10-no}.
We observe that all the clients decide to choose $\epsilon$ to be the acceptable value ($\epsilon$ = $\epsilon_{a}$) to prevent getting evicted from the learning process and obtain the global model at every round.

\begin{figure}[t]
    \centering
    \includegraphics[width=0.9\columnwidth]{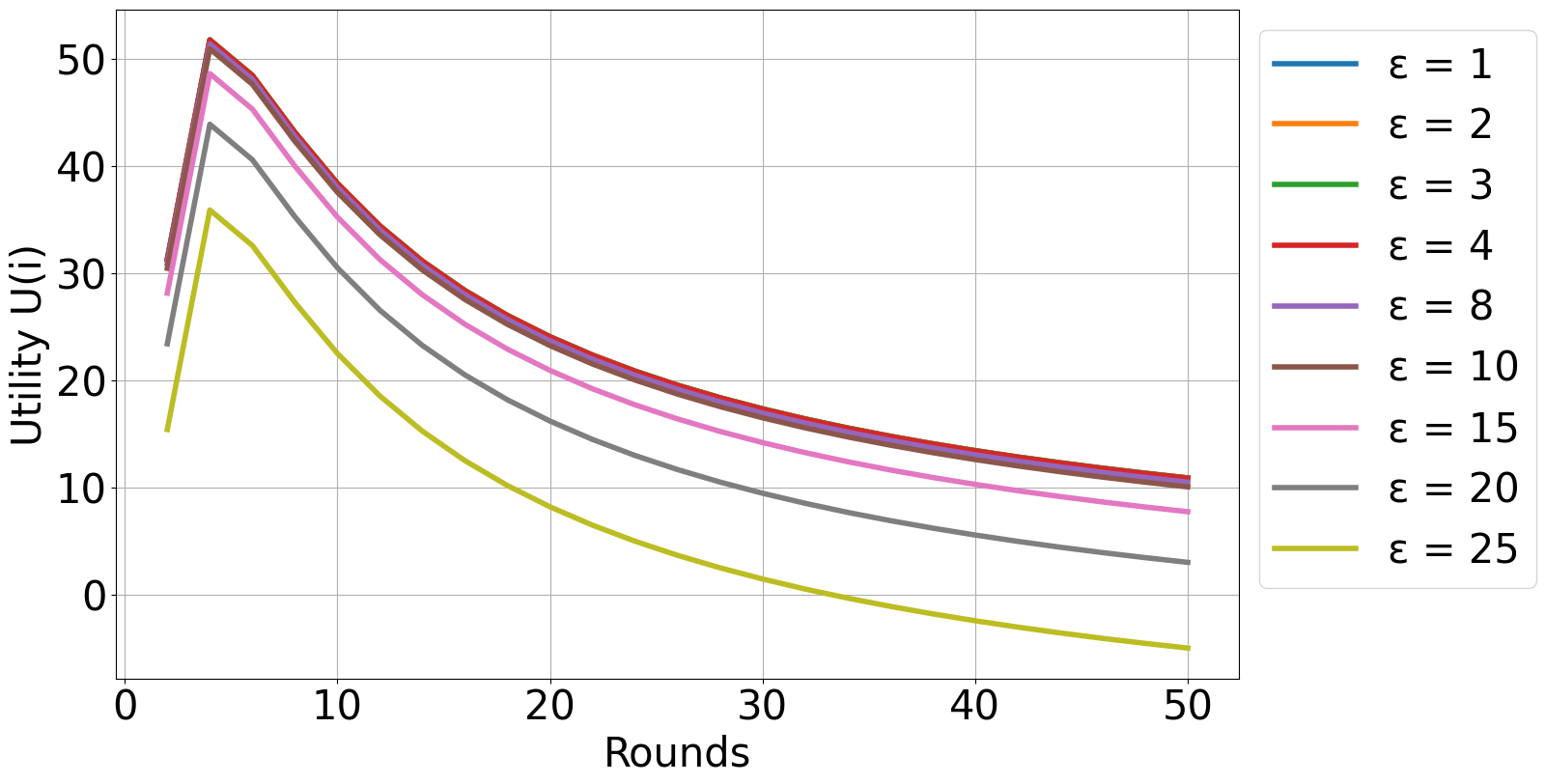}
  \setlength{\belowcaptionskip}{-8pt}
    \caption{Utility vs. No. of rounds for group-based mechanism.}
    \label{fig:utility-g}
\end{figure}

\begin{figure*}[t]
    \centering
    \captionsetup[subfigure]{}
    \subfloat[Group participation
        \label{fig:nmulg-c10}]{
\includegraphics[width=0.45\textwidth,keepaspectratio]{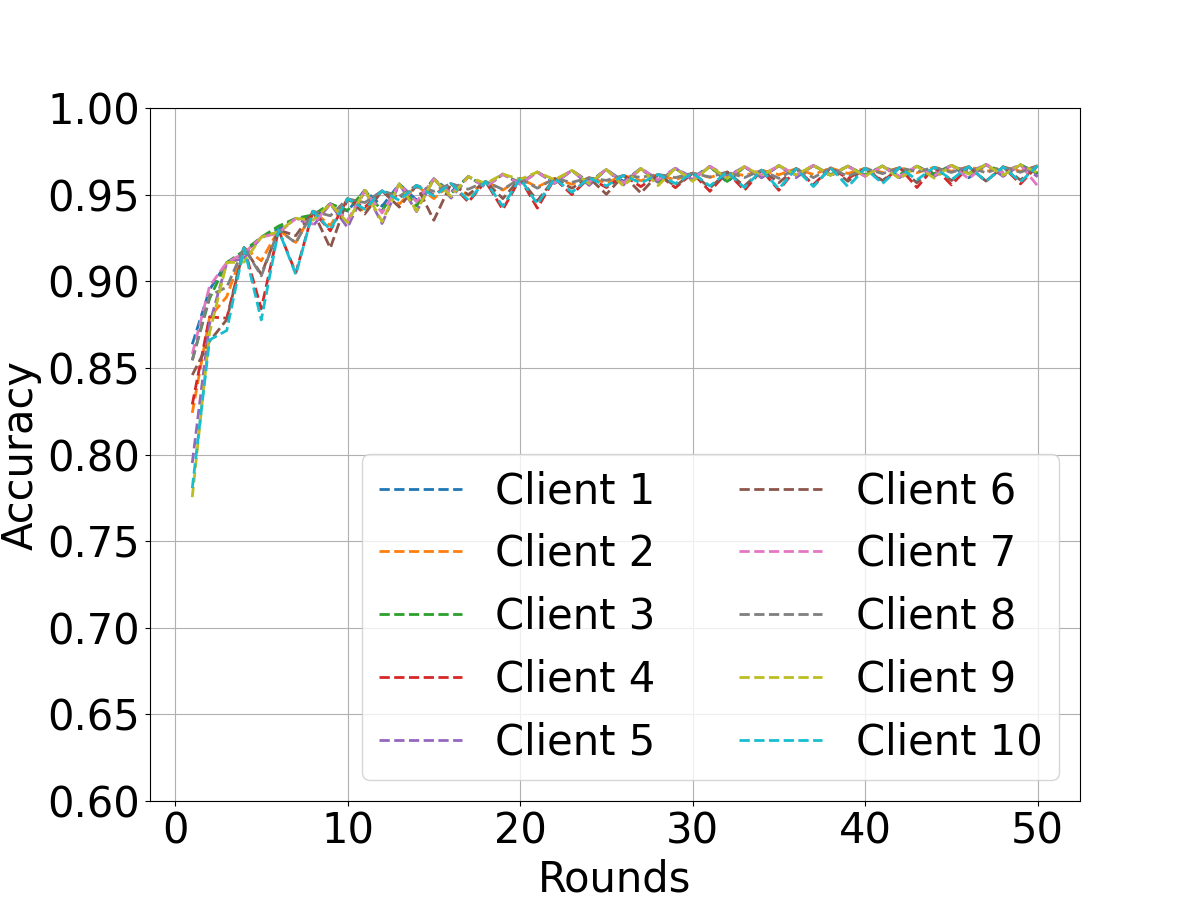}
    }
    \subfloat[Individual participation
        \label{fig:nmul-c10-e20}]{
       \includegraphics[width=0.45\textwidth,keepaspectratio]{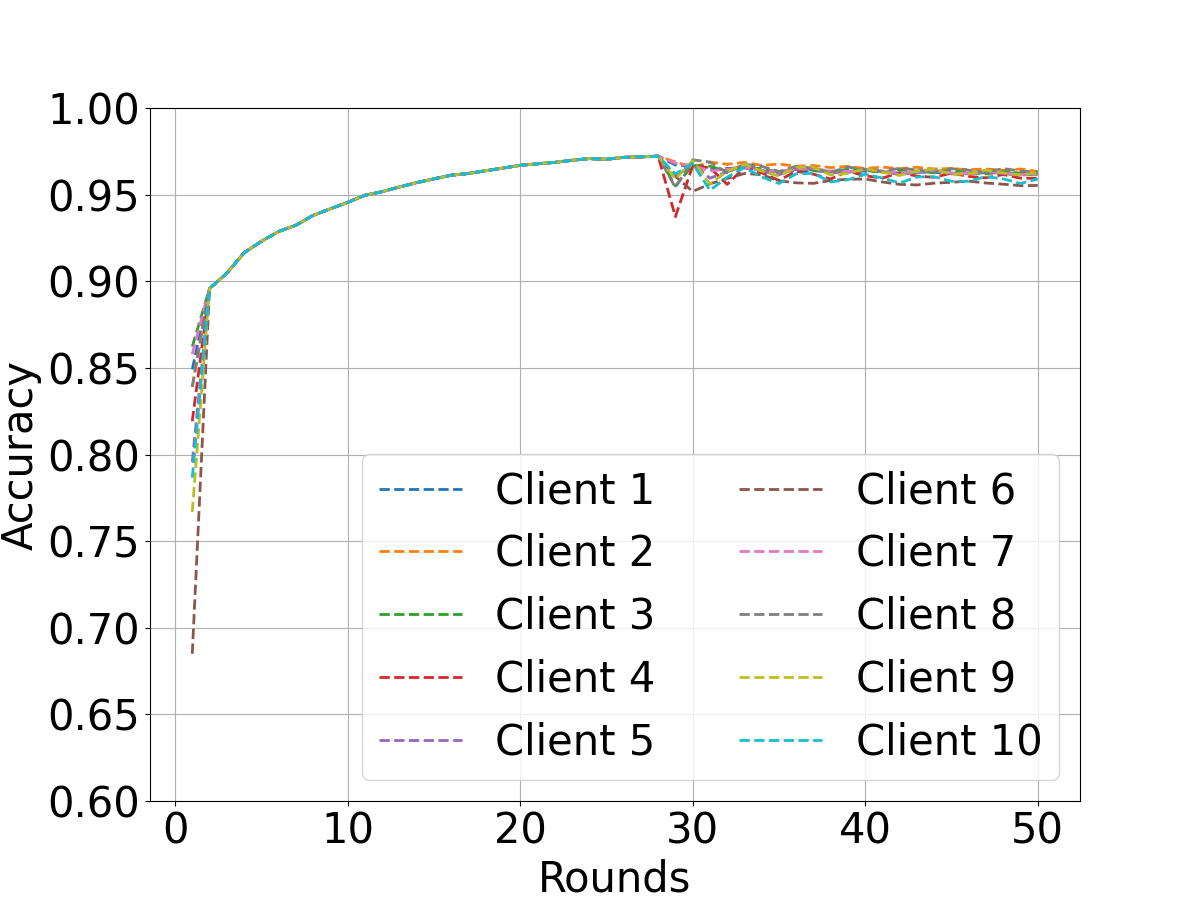}
    }
      \setlength{\belowcaptionskip}{-8pt}
    \caption{Accuracy with (a) Group participation mechanism and (b) Individual participation using \textbf{MNIST dataset}, with 10 clients with initial privacy level $\epsilon$ set to 20, and tested on a global dataset with intermediary data distribution.}
    \label{fig:groupaccuracy}
\end{figure*}

\subsection{Comparison with the Baseline SBTLF using MNIST}
\label{sec:baselinecomparison}

\textbf{Accuracy comparison}
First, we compare the results with 10 clients for both the baseline SBTLF (Figure~\ref{fig:baseline-acc-c10}) and the proposed mechanism (Figure~\ref{fig:nmul-c10-no}). In the latter, the ten accuracy lines are overlapping into one, indicating that all the clients have performed the same on the global test dataset. This means that all the clients are successfully able to receive the aggregated global model at all rounds. In contrast, in the baseline scenario, the clients with varied $\epsilon$ values obtain the global model at different intervals. This implies that in the proposed approach, the server is solving its problem of receiving high-quality model updates. The game-theoretic mechanism also solves the client's problem of being able to participate in the maximum number of rounds without getting evicted from the learning process. The carefully designed rules could capture the client-server dynamics by solving both of their problems using the proposed incentive mechanism. As an example, in Figure~\ref{fig:nmul-c10-max}, it can be seen that clients are penalized by being evicted from the learning process for not following the constraints of the privacy level to be chosen, which is not seen in the case of the baseline mechanism. Clients are not obligated to increase or decrease their privacy levels because the only penalty is not receiving the global model in certain rounds.

\textbf{Participation count comparison}
This is evident from the client participation plots of the baseline approach and the strategic mechanism illustrated in Figure~\ref{fig:baseline-pc-c10} and Figure~\ref{fig:nmul-c10-break-pc}, respectively. The baseline shows a steady decrease in the client's participation with the chosen privacy level $\epsilon$, whereas the proposed approach determines the participation based on the chosen privacy levels $\epsilon$ = (25, 17, and 15). For example, for $\epsilon$ = 25 and 17,  a client's early eviction is apparent from the blue and orange bars, whereas when $\epsilon$ = 15, all the clients show full participation.

\subsection{Results of the Mechanism with Grouping of Clients}\label{subsec:groupresults}
We next present the results of the improved mechanism using client groups, as explained in Sub-section~\ref{subsubsec:improvedmechanism}. We study the learning process with 10 clients, but instead of allowing all clients to participate in every round, we divide them into groups of 5, effectively forming a total of 2 groups. In each round, only one group is selected to participate, which follows a round-robin pattern.
The client utility values are plotted in Figure~\ref{fig:utility-g} with different $\epsilon$ values.
It is observed that, 
at $\epsilon$ = 20, unlike the individual participation case, the utility does not go below zero before the end of the learning process (Round 50). Even for $\epsilon$=25, utility becomes negative much later (instead of 18, now it is 32). 
Thus, the group-based mechanism is more robust than individual participation.

Finally, we make head-to-head comparison between the group participation and the individual participation mechanisms in Figure \ref{fig:groupaccuracy}. In this experiment with MNIST, we set the initial privacy level $\epsilon$ = 20 for all 10 clients. Figure~\ref{fig:nmulg-c10} shows the accuracy of clients participating in groups over the rounds in which they participated, continuing until the model reaches convergence.  From the plot, it can be observed that all the clients have participated in all the rounds they were chosen. In contrast, Figure~\ref{fig:nmul-c10-e20} considers that all the 10 clients are allowed to participate in every round without any group selection starting with $\epsilon$ = 20. In this process, we can see that at round 28, the training collapses. For the group-based scheme, the training does not collapse. The utility of the client does not go below zero (Figure \ref{fig:utility-g}) since the difference of the global model obtained in the current round and in the previous round the client participated in, is large enough to compensate for the cost incurred with the chosen privacy level at $\epsilon$ = 20.


\section{Related Work}
\label{sec:related}

In this section, we review existing literature on federated learning that are closely related to our work.
\subsection{Federated Learning and Privacy Concerns}
\label{subsec:relatedFLPrivacy}
Kairouz et al. \cite{adv-open-problems-fl} discuss advances and open problems in \ac{fl}, and explore critical aspects around ensuring privacy. Their work also addresses data heterogeneity in decentralized model training while highlighting some of the core issues such as communication bottleneck, client-device constraints, and model robustness under diverse data distributions. The requirement for a privacy-preserving model due to persistent security concerns and adversarial threats in \ac{fl} environments is also analyzed. Hao et al. \cite{towards-efficient-privacy-preserving-fl} demonstrate that adversaries can deduce user information from local outputs, such as gradients, and conclude that a privacy-preserving scheme is of critical importance in \ac{fl}. In response, Arachchige et al. \cite{ldp-for-fl} propose an LDP-FL protocol to introduce privacy guarantees against untrusted entities like the server, or any adversary that could compromise the server.

\subsection{Incentive Mechanisms in Federated Learning}
\label{subsec:relatedIncentive}
Most of the existing incentive mechanisms offer monetary rewards as incentives, but the ultimate goal of clients, especially in a cross-silo set up, is to receive a better-performing model. For instance, Sarikaya and Ercetin \cite{gt1} introduce a game-theoretic approach to incentivize clients in \ac{fl}, where clients are rewarded based on their contributions. Zhan et al. \cite{gt2} propose a similar mechanism in which clients are incentivized based on the quality of their contributions. In contrast, Feng et al. \cite{gt3} suggest a reward mechanism based on the privacy level of the client's data, while Zeng et al. \cite{gt4} use a token-based system to incentivize clients. Jiao et al. \cite{gt5} propose a mechanism that rewards clients based on the accuracy of their contributions, whereas Le et al. \cite{gt6} suggest a reward system based on the fairness of the client's data.

However, these mechanisms may not be universally effective, as clients are often more interested in model accuracy than monetary rewards. Towards this, Kong et al. \cite{incentivized-fl} propose a non-monetary incentive mechanism that uses model performance as reward. The server evaluates the performance of each client's uploaded model in every round and distributes different models to the clients based on the evaluation results. But as the data is heterogeneous, the model performance may not be a fair metric to evaluate the client's contributions. Lyu et al. \cite{towards-fair-privacy-preserving-fl} propose a decentralized \ac{fppdl} framework to incorporate fairness into federated deep learning models. Under \ac{fppdl}, each client receives a different version of the FL model with performance commensurate with its contributions, but the client's actions are not optimized as a game. Wu et al. \cite{incentive-aware-fl} introduce an \ac{fl} framework where the clients receive different models, each aggregated using a specific proportion of local model updates such that higher-contributing clients receive models aggregated from a larger proportion of local model updates. Yu et al. \cite{sustainable-incentive-fl} design an incentive mechanism for the clients to participate, accounting for the delays before the server could pay back the clients. It also gives five different pay-off sharing schemes for the same. Chaudhury et al., in contrast, propose a blockchain-based LDP-FL mechanism along with tokens as incentives for the clients to participate \cite{saptarshi-IEEETP}. However, this approach uses a na\"ive tokenization system where the clients can use these tokens to obtain the updated global model. 

\subsection{Game Theory in Federated Learning}
Game theory frameworks can be applied to model strategic interactions between clients and servers in FL, one of the most common being the Stackelberg game \cite{note-on-stackelberg-gt}. It is a decisive game between a leader and followers, where the leader makes the first move and the followers respond. This interaction is strategically modeled to optimize both the leader's and the followers' objectives. Xu et al. 
\cite{incentive-private-fl-iot} apply Stackelberg game theory in FL in which the server acts as the leader and the clients as followers. Donahue and Kleinberg
\cite{optimality-stability-fl-gt} introduce cooperative games for balancing client incentives, but often assumes static client strategies. Rathi et al. 
\cite{stackelberg-gt-d2d} also explore a leader-follower dynamics. However, there framework lacks mechanisms for handling privacy-accuracy trade-offs unique to FL.
He et al. \cite{gt-incentive-fl-blockchain} propose incentive mechanisms using game theory on blockchain-based FL. Witt et al. \cite{decentral-incentive-fl-lr} perform a systematic literature review on decentralized incentive mechanisms in \ac{fl}.
Several other studies such as \cite{fl-gt-secure-iiot}, \cite{incentive-private-fl-iot}, \cite{gt-fl-privacy-accuracy-energy}, \cite{gt1}, \cite{gt-robust-fl}, \cite{colab-fl-gt} also apply game theory to FL, but they do not address the privacy-accuracy trade-offs in client strategies.

\section{Conclusion}
\label{sec:concl}

In this paper, we have proposed a game theoretic formulation of incentivized LDP-FL systems in a cross-silo FL setup. Tokens are awarded to participating clients based on the level of privacy chosen by them in any given round. Such acquired tokens are later required to obtain the updated global models from the server. Different incentive mechanisms have been designed and their corresponding strategic decisions were discussed. Extensive experiments were carried out to study the impact of various parameters on the accuracy of the model training process and the degree of client participation using the MNIST dataset. Experiments on the proposed game-theoretic incentive mechanisms were also performed on the CIFAR10 dataset to prove that the results are not dependent on the dataset used. It is shown that strategic incentivization results in significant improvement over the baseline approach.

Future work involves exploring other incentive mechanisms and their strategic analysis. Use of larger datasets and different machine learning problems can also be looked into. A challenging new direction would be to consider the impact of adversarial attacks on the proposed scheme.

\begin{figure*}
  \centering
  \captionsetup[subfigure]{}
  \subfloat[$\epsilon = 25$\label{fig:nmul-c3-max-cifar}]{
    \includegraphics[width=0.32\textwidth,keepaspectratio]{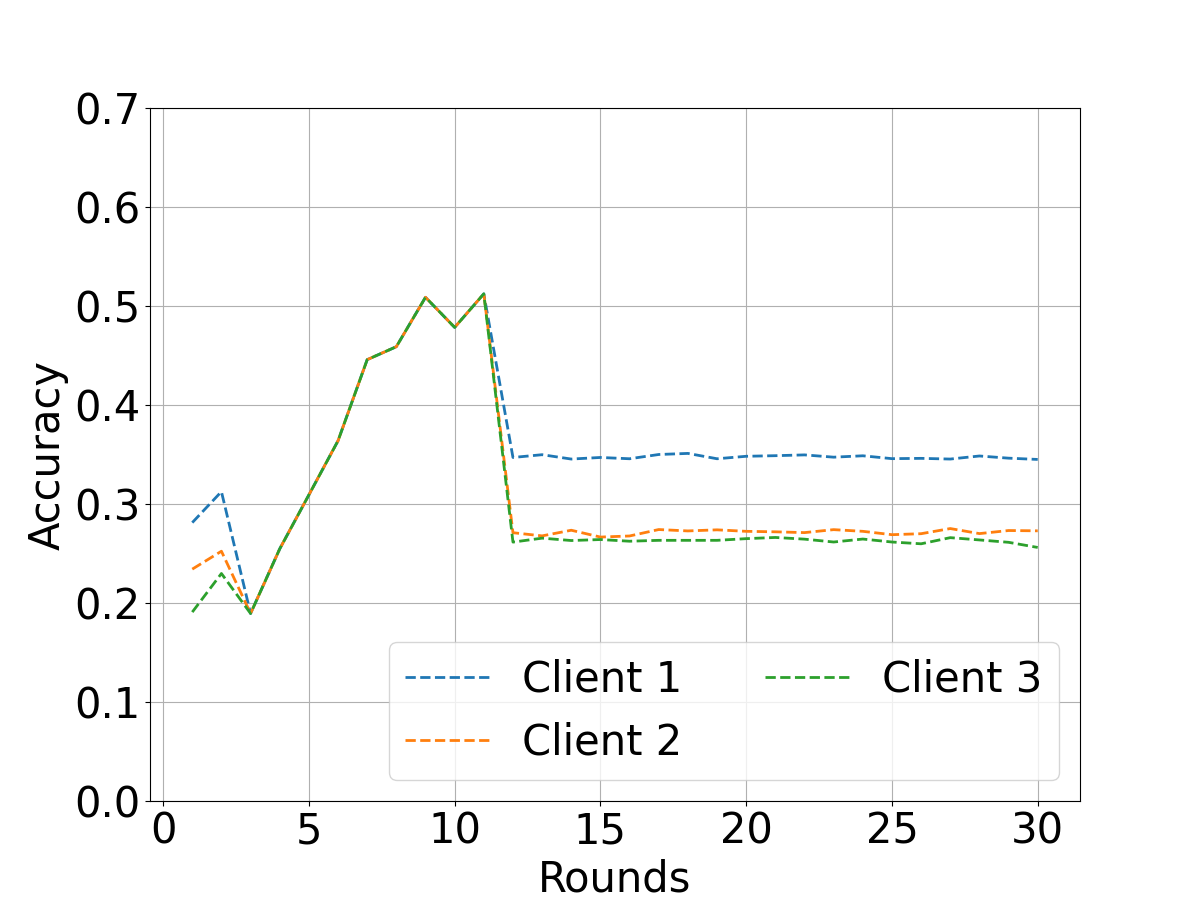}
  }
  \subfloat[$\epsilon = 20$\label{fig:nmul-c3-min-cifar}]{
    \includegraphics[width=0.32\textwidth,keepaspectratio]{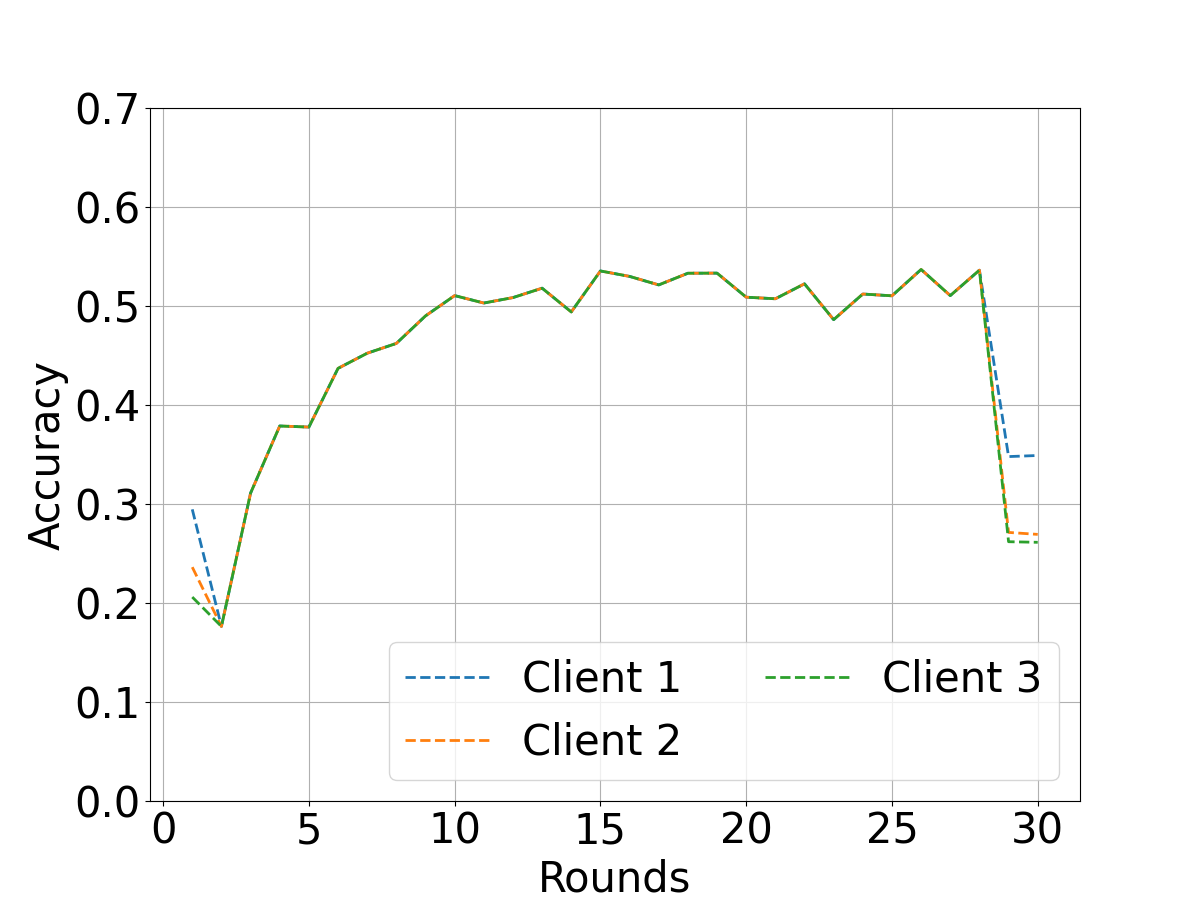}
  }
  \subfloat[$\epsilon = 15$\label{fig:nmul-c3-no-cifar}]{
    \includegraphics[width=0.32\textwidth,keepaspectratio]{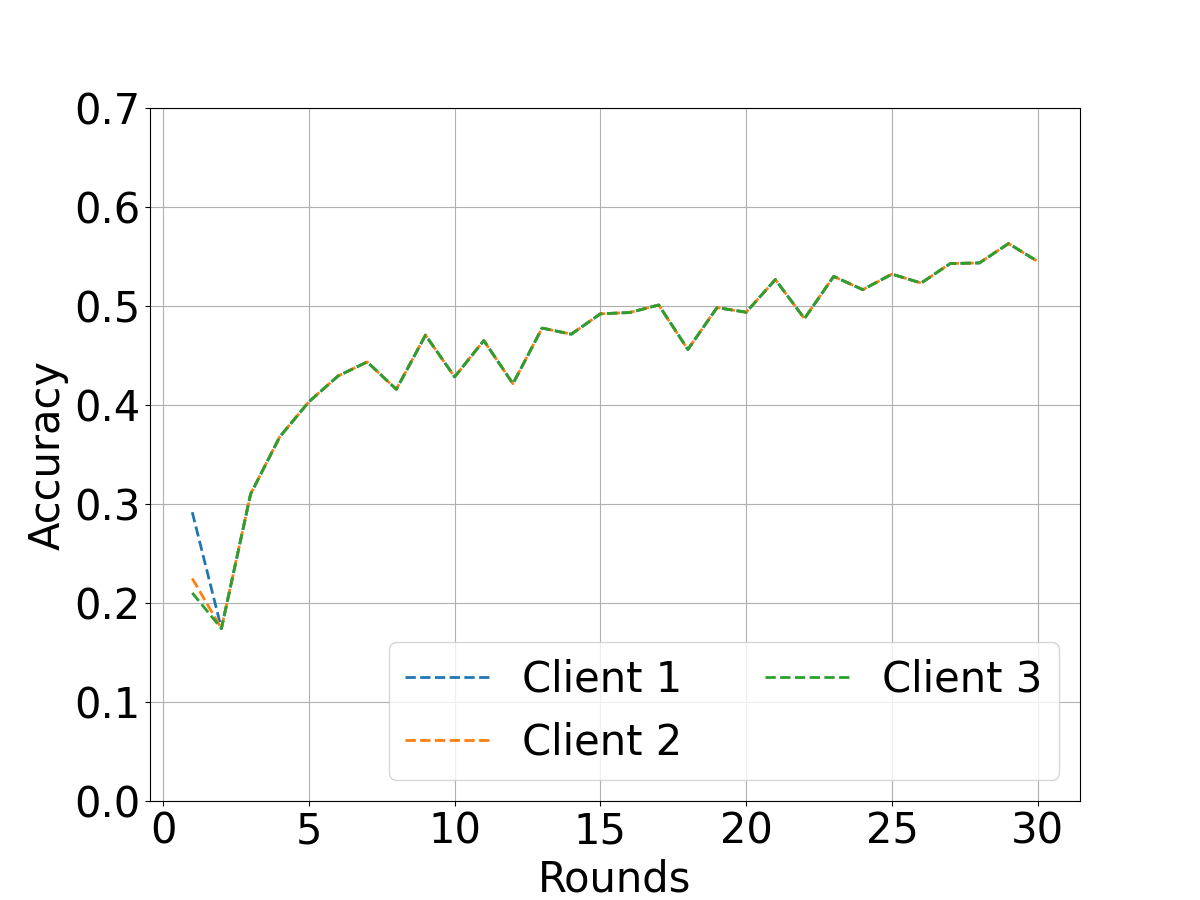}
  }
  \setlength{\belowcaptionskip}{-8pt}
    \caption{FL Accuracy variation for the \textbf{CIFAR10 dataset} tested on a Global Dataset with Disjoint Distribution for \textbf{3} clients illustrating where the training collapse occurs (a) earliest, (b) after most delay, and (c) does not collapse at all.}
  \label{fig:nmul-c3-breakpoint-cifar}
\end{figure*}

\begin{figure*}
  \centering
  \captionsetup[subfigure]{}
  \subfloat[$\epsilon = 25$
    \label{fig:nmul-c10-max-cifar}]{
    \includegraphics[width=0.32\textwidth,keepaspectratio]{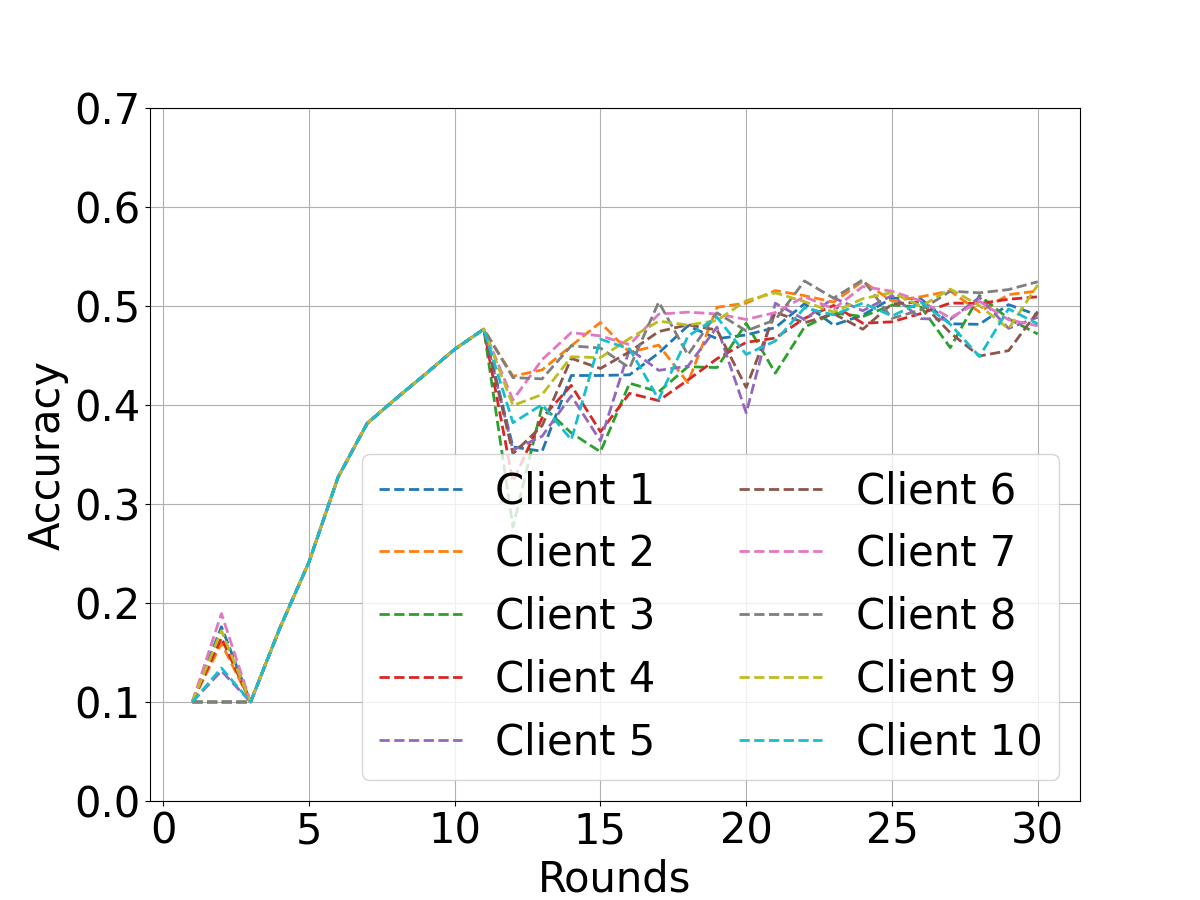}
  }
  \subfloat[$\epsilon = 20$
    \label{fig:nmul-c10-min-cifar}]{
    \includegraphics[width=0.32\textwidth,keepaspectratio]{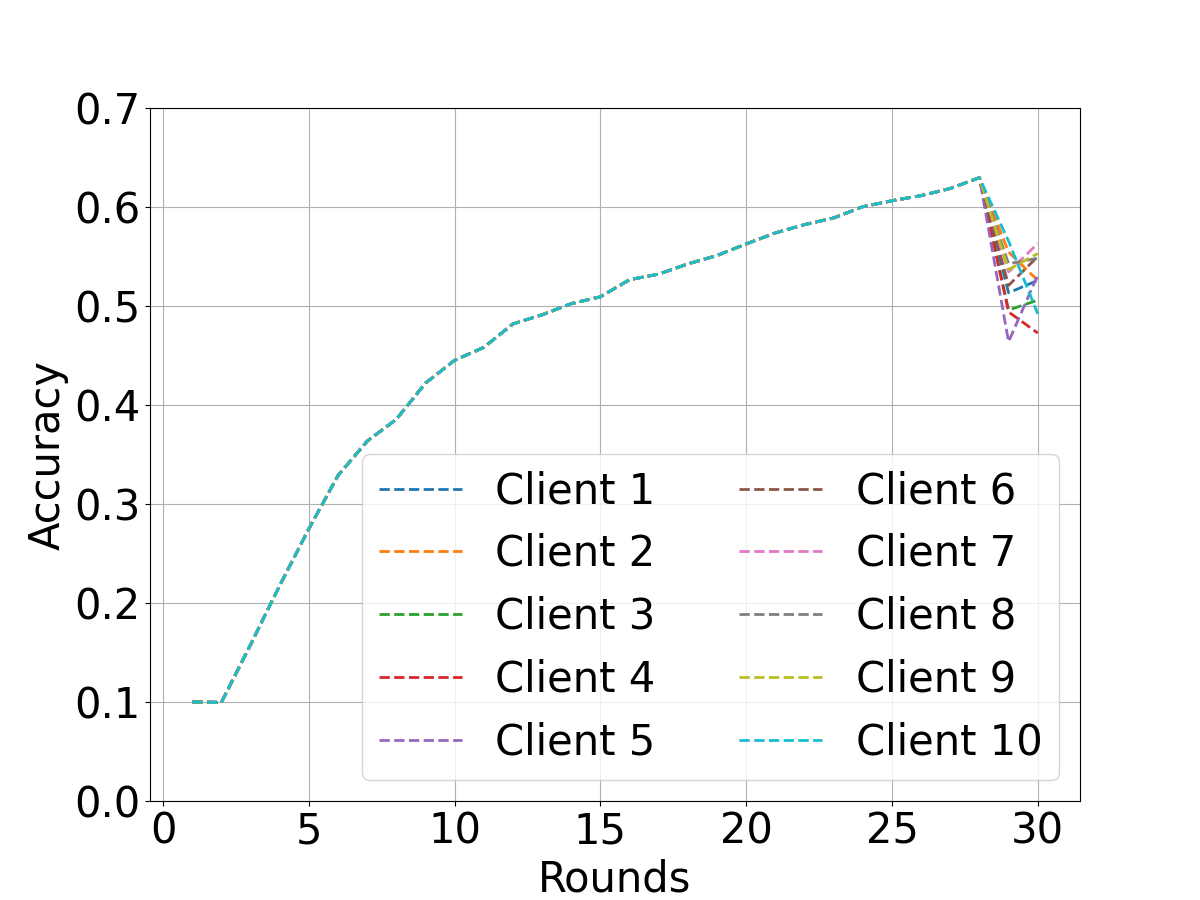}
  }
  \subfloat[$\epsilon = 15$
    \label{fig:nmul-c10-no-cifar}]{
    \includegraphics[width=0.32\textwidth,keepaspectratio]{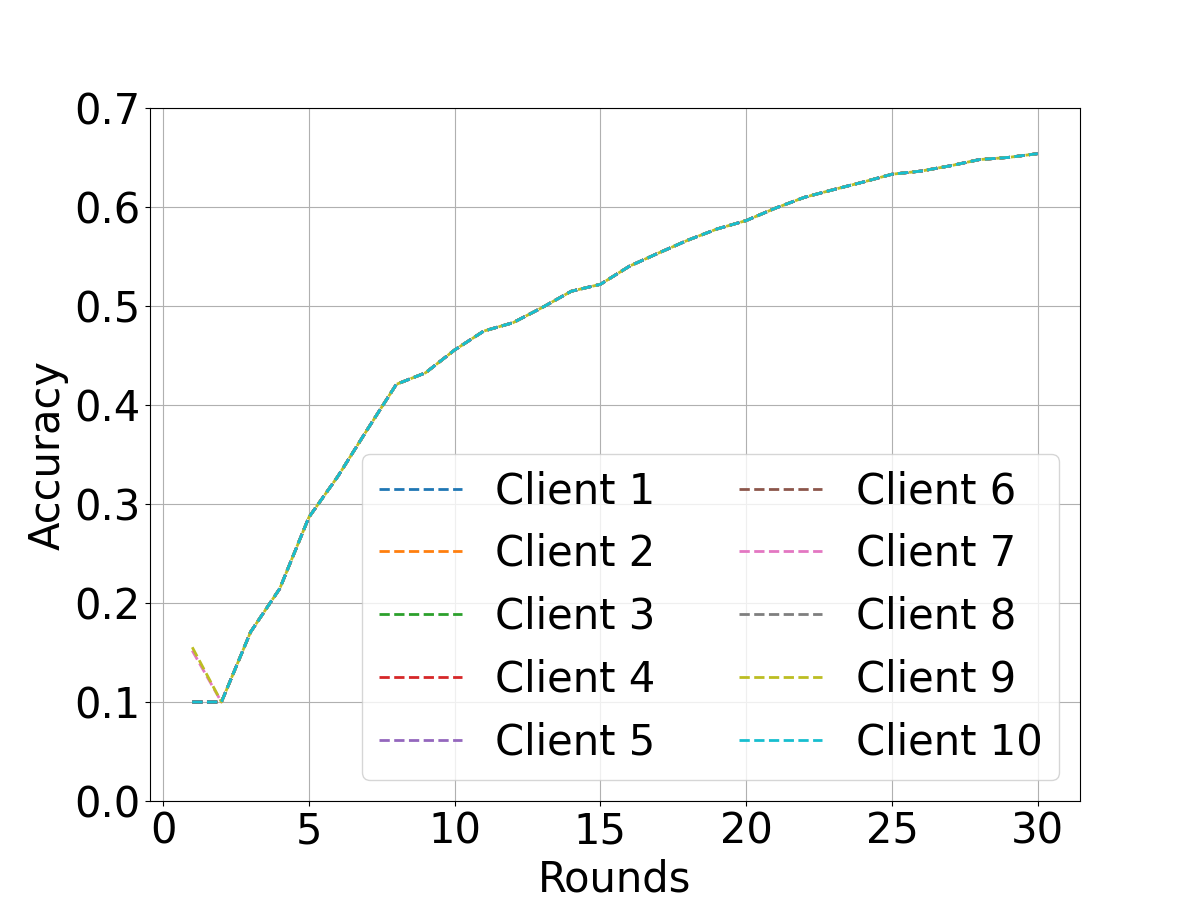}
  }
  \setlength{\belowcaptionskip}{-8pt}
      \caption{FL Accuracy variation for the \textbf{CIFAR10 dataset} tested on a Global Dataset with Disjoint Distribution for \textbf{10} clients illustrating where the training collapse occurs (a) earliest, (b) after most delay, and (c) does not collapse at all.}
  \label{fig:nmul-c10-breakpoint-cifar}
\end{figure*}

\section*{Acknowledgments}
The research work reported in this paper was supported partially by CISCO University Research Program Fund, Silicon Valley Community Foundation award number 2020-220329 (3696) and partially by Rutgers Business School Dean’s Research Seed Fund awarded in 2025.


\bibliographystyle{IEEEtran}
\bibliography{sample-base}

\appendix
\label{sec:appendix}
For the sake of brevity, the following figures on experimental results using the CIFAR10 dataset were not included in the main text and hence, those are being presented here.

\begin{figure*}[]
  \centering
    \captionsetup[subfigure]{}
  \subfloat[3 Clients\label{fig:nmul-c3-break-pc-cifar}]{
    \includegraphics[width=0.45\textwidth]{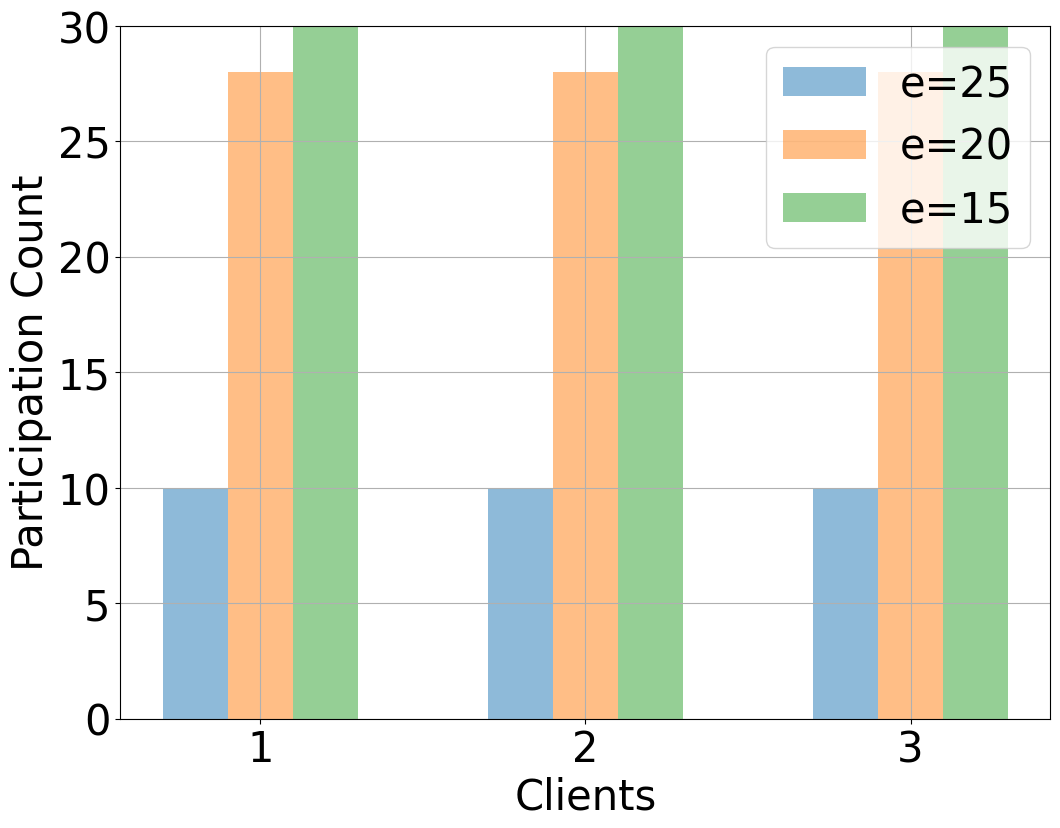}
  }
  \subfloat[10 Clients\label{fig:nmul-c10-break-pc-cifar}]{
    \includegraphics[width=0.45\textwidth]{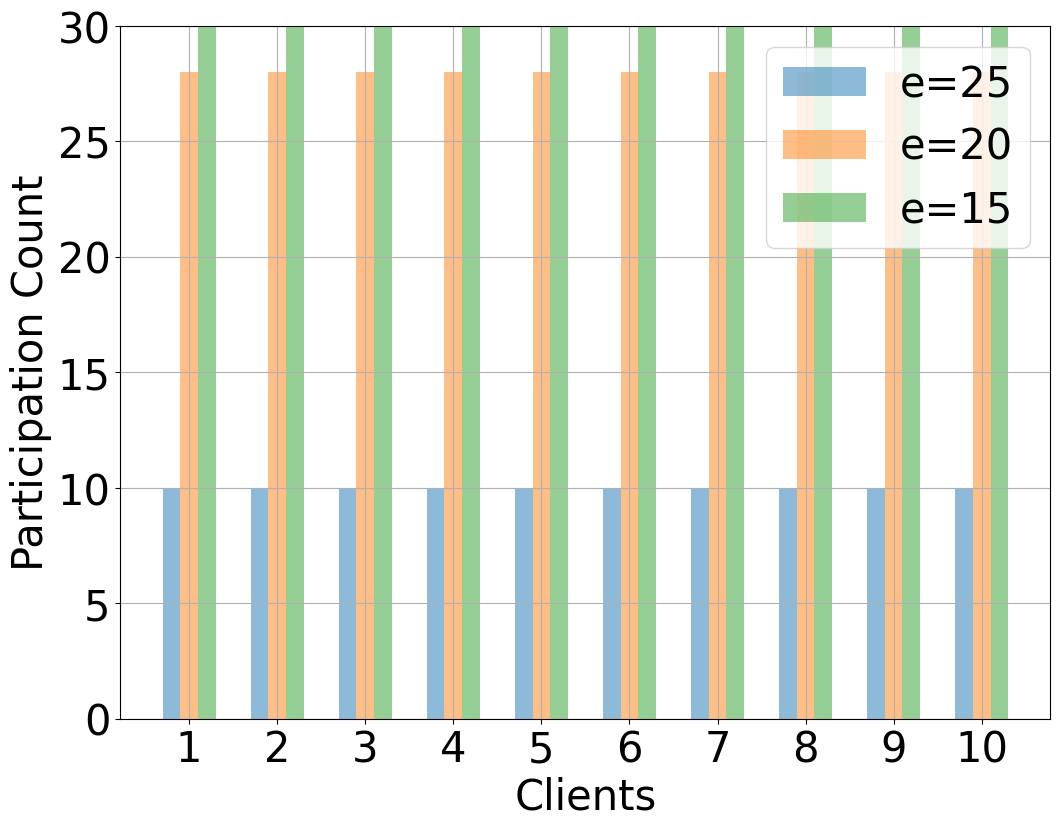}
  }
      \setlength{\belowcaptionskip}{-8pt}
  \caption{Client participation count tested on a Global Dataset using the \textbf{CIFAR10 dataset}. The figure shows participation across (a) 3 and (b) 10 clients. Each bar indicates how often a client participated when training collapsed early (blue), after a long delay (orange), or never (green).}
  \label{fig:nmul-breakpoint-pc-cifar}
\end{figure*}

\begin{figure*}[t]
    \centering
    \captionsetup[subfigure]{}
    \subfloat[Group participation
        \label{fig:nmulg-c10-cifar}]{
        \includegraphics[width=0.45\textwidth,keepaspectratio]{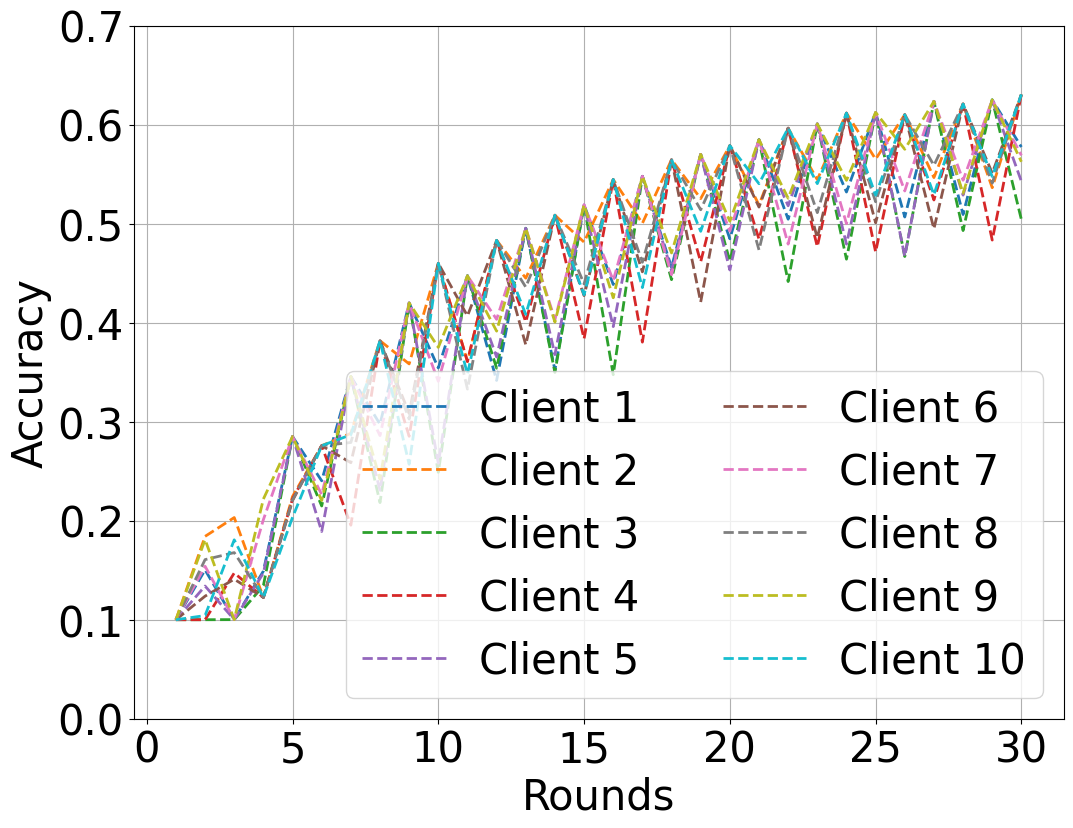}
    }
    \subfloat[Individual participation
        \label{fig:nmul-c10-e25-cifar}]{
\includegraphics[width=0.51\textwidth,keepaspectratio]{cifaroutput/clients10/nmul/intermediary/global_e25_r30_fl1_multi0.png}
    }
    \caption{Accuracy with (a) Group participation mechanism and (b) Individual participation using \textbf{CIFAR10 dataset}, with 10 clients with initial privacy level $\epsilon$ set to 25, and tested on a global dataset with intermediary data distribution.}
    \label{fig:groupaccuracy-cifar}
\end{figure*}

CIFAR10 consists of 60,000 color images in 10 classes, with 6,000 images per class. The dataset is divided into 50,000 training images and 10,000 test images. Each image is of size 32$\times$32 pixels. Unlike MNIST, we implemented a \ac{cnn} model suitable for classifying the CIFAR10 dataset. The model architecture was initialized as follows:
\begin{enumerate}
    \item A 2D convolutional layer with 32 filters (3$\times$3), ReLU activation, followed by a 2D max pooling layer (2$\times$2).
    \item Next, a convolutional layer with 64 filters (3$\times$3), ReLU activation, followed by 2D max pooling layer (2$\times$2).
    \item A third convolutional layer with 128 filters (3$\times$3) and ReLU activation.
    \item A flatten layer followed by a dense layer with 128 neurons and ReLU activation followed by an output dense layer with 10 units (no activation).
\end{enumerate}
For the baseline scheme, we considered only use the MNIST dataset, while for the proposed schemes in the paper, we use both the MNIST and CIFAR10 datasets.
In subsection~\ref{subsec:strategicresults}, we detailed the observations for experiments conducted using the MNIST dataset. Here, we provide observations for the experiments conducted using the CIFAR10 dataset.
The utility function for CIFAR10 is the same as that of MNIST, but the number of rounds is lesser. The observations are similar to those of MNIST, except that the utility remains positive for a longer duration. For $\epsilon = 20$, unlike MNIST, the utility drops below zero at the very end of the training process. The cases for $\epsilon > 20$ and $\epsilon < 15$ are the same as those of the training using MNIST. For $\epsilon < 20$, the utility remains positive throughout the training.

We also examine at which $\epsilon$ values the training collapses with varying number of clients: 3 (Figure~\ref{fig:nmul-c3-breakpoint-cifar}(a)-(c)) and 10 (Figure~\ref{fig:nmul-c10-breakpoint-cifar}(a)-(c)). These figures present the accuracy graph plotted against rounds with the participating clients of 3 and 10, respectively using the CIFAR10 dataset.
It is evident that the utility function is dataset independent, but the difference in the number of rounds in the learning process (MNIST: 50, CIFAR10: 30) is responsible for the $\epsilon$ values observed above to be different. The $\epsilon$ values for the three cases are as follows:
\begin{itemize}
  \item MNIST: (a) $\epsilon$ = 25, (b) $\epsilon$ = 17, (c) $\epsilon$ = 15
  \item CIFAR10: (a) $\epsilon$ = 25, (b) $\epsilon$ = 20, (c) $\epsilon$ = 15
\end{itemize}.

The clients' participation behavior for CIFAR10 is similar to that of MNIST, which was discussed in its respective subsection.
The participation count plots in Figure~\ref{fig:nmul-c3-break-pc-cifar} and Figure~\ref{fig:nmul-c10-break-pc-cifar} show how many training rounds each client participated in for the three-client case (Figure~\ref{fig:nmul-c3-break-pc-cifar}) and the ten-client case (Figure~\ref{fig:nmul-c10-break-pc-cifar}) under different values of $\epsilon$ (25, 20, 15). These trends align with the accuracy behaviors observed earlier.

We next consider the improved mechanism with grouping of clients with CIFAR10. We set the initial privacy level $\epsilon$ = 25. Recall that the utility remains positive until the end of 30 rounds. Figure~\ref{fig:groupaccuracy-cifar} shows the accuracy of clients participating in the experiment same as in Figure~\ref{fig:nmulg-c10} but with CIFAR10 dataset. The accuracy of the clients is shown for the group-based mechanism in Figure~\ref{fig:nmulg-c10-cifar}. It can be observed that the accuracy of the clients is much lower than that of the MNIST dataset, but it is still higher than the individual participation case (Figure~\ref{fig:nmul-c10-e25-cifar}). The utility remains positive until round 30, which is the end of the learning process.



\end{document}